\documentclass{opendatalab}

\usepackage[utf8]{inputenc}
\usepackage{xcolor}
\usepackage[most]{tcolorbox}
\usepackage{longtable}
\usepackage{listings}
\usepackage[numbers]{natbib}
\usepackage{enumitem}
\usepackage{graphicx}
\usepackage{xspace} 
\usepackage{hyperref}
\usepackage[table]{xcolor}    % 用于单元格背景色渲染
\usepackage{pdflscape}        % 用于局部页面横置
\usepackage{booktabs}         % 提供三线表支持（tabular/tabularx 的标配）
\usepackage{tabularx}         % 自动调整列宽以适应页面
\usepackage[T1]{fontenc}
\usepackage{url}            % simple URL typesetting
\usepackage{nicefrac}       % compact symbols for 1/2, etc.
\usepackage{microtype}      % microtypography
\usepackage{graphicx}
\definecolor{codegreen}{rgb}{0,0.6,0}
\definecolor{codegray}{rgb}{0.5,0.5,0.5}
\definecolor{codepurple}{rgb}{0.58,0,0.82}
\definecolor{backcolour}{rgb}{0.95,0.95,0.92}
\definecolor{promptcolor}{HTML}{D1D0F2}
\definecolor{promptcolorheader}{HTML}{bdbcec}

\newcommand{\github}{\raisebox{-1.5pt}{\includegraphics[height=1.05em]{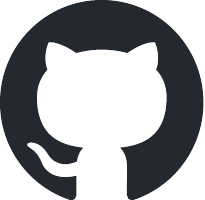}}\xspace}
\newcommand{\web}{\raisebox{-1.5pt}{\includegraphics[height=1.05em]{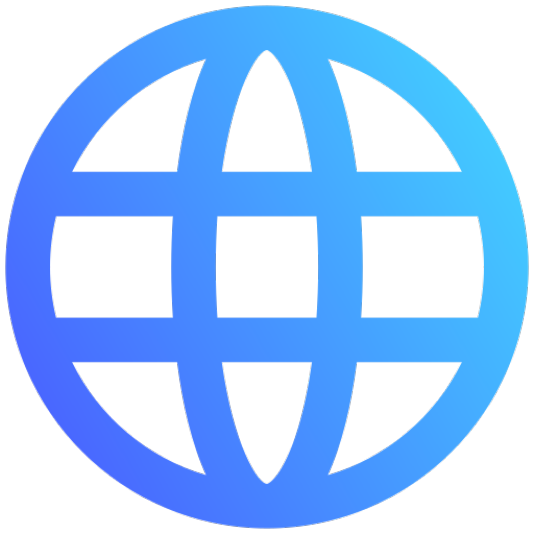}}\xspace}
\newcommand{\huggingface}{\raisebox{-1.5pt}{\includegraphics[height=1.05em]{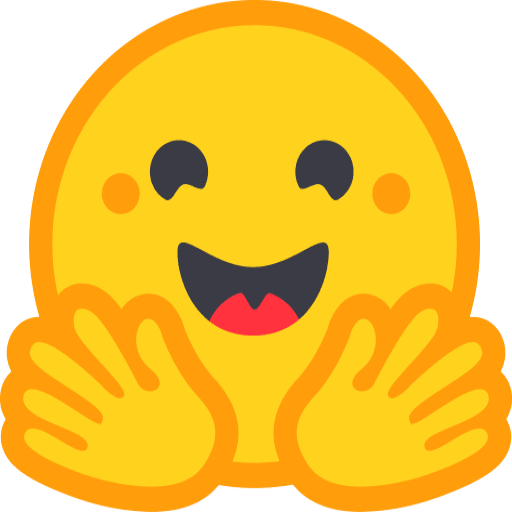}}\xspace}

\definecolor{promptcolor}{HTML}{EAF3FF}
\definecolor{promptcolorheader}{HTML}{8DB7E8}
\definecolor{codeback}{HTML}{F7FAFF}
\newtcolorbox{promptbox}[1][]{
  enhanced,
  breakable,
  top=0.3em,
  bottom=0.3em,
  left=0.5em,
  right=0.5em,
  toptitle=0.3em,
  bottomtitle=0.2em,
  boxsep=0pt,
  colframe=promptcolorheader,
  colback=promptcolor!50,
  boxrule=0.5pt,
  width=\columnwidth,
  title={\footnotesize #1}
}
\title{\textsc{PAGER}: Bridging the Semantic-Execution Gap in Point-Precise Geometric GUI Control}

\author[1*]{Jingxuan Wei}
\author[1*]{Xi Bai}
\author[3*]{Shan Liu}
\author[1*]{Caijun Jia}
\author[1]{Zheng Sun}
\author[1]{Xinglong Xu}
\author[2]{Siyuan Li}
\author[1]{Linzhuang Sun}
\author[1]{ Bihui Yu}
\author[2]{Conghui He}
\author[2]{Cheng Tan}

\affiliation[1]{University of Chinese Academy of Sciences}
\affiliation[2]{Shanghai Artificial Intelligence Laboratory}
\affiliation[3]{China University of Petroleum-Beijing}

\abstract{
Large vision-language models have significantly advanced GUI agents, enabling executable interaction across web, mobile, and desktop interfaces.
Yet these gains largely rely on a forgiving \emph{region-tolerant} paradigm, where many nearby pixels inside the same component remain valid.
Precise geometric construction breaks this assumption: actions must land on points in continuous canvas space rather than tolerant regions.
Because geometric primitives carry ontological dependencies, a local coordinate error can induce cascading topological failures that distort downstream objects and invalidate the final construction.
We identify this regime as \emph{precision-sensitive GUI tasks}, requiring point-level accuracy, geometry-aware verification, and robustness to dependency-driven error propagation.
To benchmark it, we introduce \textsc{PAGE} Bench, with 4{,}906 problems and over 224K process-supervised, pixel-level GUI actions.
We further propose \textsc{PAGER}, a topology-aware agent that decomposes construction into dependency-structured planning and pixel-level execution.
Pixel-grounded supervised tuning establishes executable action grammar, while precision-aligned reinforcement learning mitigates rollout-induced exposure bias through state-conditioned geometric feedback.
Experiments reveal a pronounced \textit{Semantic-Execution Gap}: general multimodal models can exceed 88\% action type accuracy yet remain below 6\% task success.
\textsc{PAGER} closes this gap, delivering $4.1\times$ higher task success than the strongest evaluated general baseline and raising step success rate from below 9\% for GUI-specialized agents to over 62\%, establishing a new state of the art for point-precise GUI control.
}

\date{\today}
\correspondence{Cheng Tan, \email{tancheng@pjlab.org.cn}
\\
\\
  \parbox{\linewidth}{\centering
    \github~\href{https://github.com/OpenRaiser/Pager}{\textbf{Code}} \quad
    \web~\href{https://openraiser.github.io/Pager-webpage/}{\textbf{Website}} \quad
\huggingface~\href{https://huggingface.co/datasets/OpenRaiser/Pager}{\textbf{Dataset}}
  }
}

\begin{document}

\maketitle

% \tableofcontents

\section{Introduction}
\label{sec:introduction}

Modern GUI agents increasingly turn software interfaces into action spaces for vision-language models. Recent systems operate across web, mobile, desktop, and broader computer-use environments by grounding multimodal instructions to interface elements and composing them into executable workflows~\cite{nguyen2025gui,chen2025guicourse,qin2025uitars,liu2026infiguiagent,zhao2025worldgui,yang2026probench,wang2026history}. This progress, however, is built mainly on a region-tolerant interaction paradigm: a button, link, input box, or menu item remains correct under many nearby click locations. The paradigm supports much of today's GUI automation, but it leaves a basic capability boundary unresolved: can an agent still operate reliably when the target is a point in continuous visual space rather than a tolerant region?

\begin{figure*}[t]
    \centering
    \includegraphics[width=\linewidth]{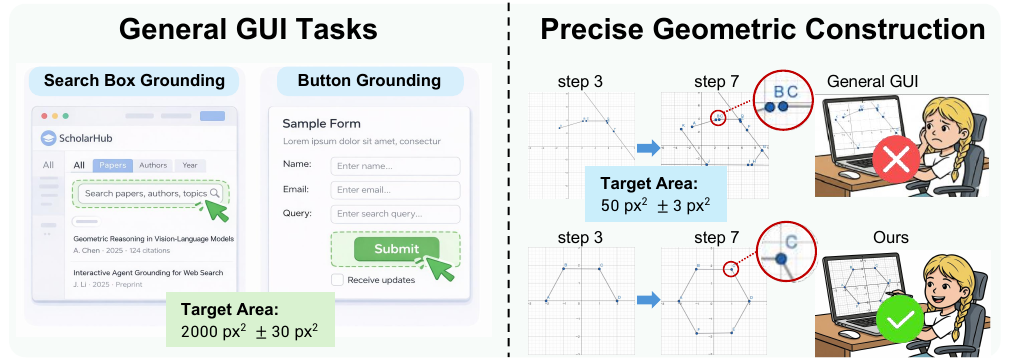}
    \caption{
    Precision-sensitive GUI tasks expose a capability gap hidden by conventional GUI benchmarks.
    In region-tolerant interaction, nearby pixels inside the same interface component lead to the same state transition.
    In precise geometric construction, an action targets a point on a continuous canvas; small coordinate errors alter geometric constraints and propagate through dependent objects.
    % \textsc{PAGE} Bench evaluates this regime, and \textsc{PAGER} addresses it through dependency-structured planning and precision-aligned execution.
    }
    \vspace{-6mm}
    \label{fig:intro_motivation}
\end{figure*}

We investigate this boundary through precise geometric construction. Given a geometry problem, the agent must construct points, segments, lines, circles, polygons, labels, and spatial relations on a GUI canvas. As illustrated in Figure~\ref{fig:intro_motivation}, this setting is not merely a harder instance of GUI grounding; it changes the success geometry from region membership to point-level accuracy within a small pixel tolerance. More importantly, geometric operations are dependency-coupled: a misplaced point changes every line, circle, intersection, angle, or polygon that depends on it, so local coordinate errors propagate through the construction process like perturbations under a dependency Jacobian. 

We call this regime \emph{precision-sensitive GUI tasks}, where agents must move beyond region-level component selection toward point-precise manipulation. This regime sits at the intersection of GUI agents, geometric reasoning, and reinforcement learning, but none of these lines directly captures it. GUI agents mainly study semantic component grounding and workflow completion~\cite{lian2025uiagile,lee2025reguide,zhou2025guig1,liao2025beyondclicking}; geometric reasoning methods focus on diagram understanding, auxiliary construction, or formal validity in symbolic spaces~\cite{xu2025geosense,feng2025geobench,weng2025geosketch,wei2025geointr1}; and RL-based agents typically optimize discrete success, milestones, or target regions rather than continuous geometric precision~\cite{zhang2025r1vl,shi2025mobileguirl,xu2025mobilerl,lu2025uir1,xia2025guir1}. To make this missing capability measurable, we introduce \textsc{PAGE} Bench, a Precision-Aware GEometric GUI Benchmark for precise geometric construction. \textsc{PAGE} Bench contains 4{,}906 geometry problems, 53{,}277 high-level construction tasks, and 224{,}497 low-level GUI actions, with trajectories that preserve problem statements, ordered sub-tasks, canvas states, execution feedback, and pixel-level geometric annotations. Its evaluation therefore goes beyond final visual similarity, measuring process correctness, parameter precision, and final geometric validity.

We further propose \textsc{PAGER}, a \textbf{P}recision-\textbf{A}ware \textbf{GE}ometric \textbf{R}easoning framework for precision-sensitive GUI tasks. \textsc{PAGER} factorizes drawing into dependency-structured planning and pixel-level execution: the planner induces a construction graph and produces a topologically valid sub-task order, while the executor grounds each sub-task into concrete GUI actions conditioned on the current canvas state. Pixel-grounded supervised tuning first establishes executable action grammar and sequential drawing behavior. Since this imitation stage is teacher-forced, inference still suffers from exposure bias: small deviations move the rollout away from reference canvas states and can be amplified by downstream geometric dependencies. Precision-aligned reinforcement learning then optimizes action-type correctness, parameter accuracy, and rendered geometric validity, directly targeting the point-level bottleneck exposed by precision-sensitive drawing.

Experiments show that this task exposes a structural mismatch in existing agents. Strong general multimodal models often understand the intended operation, but fail to maintain the continuous parameters needed for a valid construction. Ablations further show that pixel-grounded SFT provides the execution prior, parameter-accuracy rewards drive continuous-space control, and combining action-type and parameter rewards yields the strongest task-level performance.

Our main contributions are as follows:
\begin{itemize}[leftmargin=2em]
    \vspace{-1mm}
    \item We identify and formalize \emph{precision-sensitive GUI tasks}, a class of GUI tasks that require point-level spatial accuracy, continuous-canvas manipulation, geometry-aware verification, and mitigation of cascading coordinate errors.
    \vspace{-1mm}
    \item We introduce \textsc{PAGE} Bench, to the best of our knowledge the first benchmark for evaluating GUI agents on precise geometric construction, with process-supervised trajectories, pixel-level annotations, and both process-level and final-result metrics.
    \vspace{-1mm}
    \item We propose \textsc{PAGER}, a dependency-structured planning and pixel-level execution framework trained with pixel-grounded supervised tuning and precision-aligned reinforcement learning. Experiments show that \textsc{PAGER} substantially improves precise geometric GUI execution over general VLMs and GUI-specialized agents.
\end{itemize}
\section{Related Work}

\paragraph{GUI Agents}
GUI agent research maps multimodal instructions to executable actions across web, mobile, desktop, and broader computer-use environments. Early work builds perceptual-action abstractions: CogAgent~\cite{hong2024cogagent} improves high-resolution interface understanding, while CoCo-Agent~\cite{ma2024agent} structures mobile action prediction through environment perception and conditional decomposition. Recent systems such as UI-TARS~\cite{qin2025uitars} and GUI-Libra~\cite{yang2026guilibra} move toward native end-to-end execution with reasoning-aware action modeling. A related thread improves grounding accuracy and data efficiency through continuous-reward optimization, self-evolutionary reinforcement learning, spatial reasoning, test-time search, and difficulty-aware reward correction~\cite{lian2025uiagile,yuan2025segui,lee2025reguide,zhou2025guig1}; this trajectory also extends beyond clicking to text dragging~\cite{liao2025beyondclicking}. Benchmarking likewise shifts toward realistic and process-aware evaluation, including arbitrary-state desktop automation, broader computer-use, tool-use, and browsing settings~\cite{zhao2025worldgui,yang2026probench,mu2025gui360,fan2025mcptoolbenchpp,wei2025browsecomp}. Despite this progress, existing GUI agents mainly target semantic interface elements or tolerant regions, where success depends on component selection or workflow completion. Our work instead studies canvas-based precision-sensitive GUI tasks, where success requires point-level spatial accuracy, geometric validity, and mitigation of cascading error propagation induced by small coordinate deviations.

\paragraph{Geometric Reasoning}
Geometric reasoning studies how models interpret diagrams, identify principles, and derive mathematically valid solutions from multimodal inputs. Diagnostic benchmarks analyze failures in principle identification, principle application, perception, planning, theorem use, and reflection~\cite{xu2025geosense,feng2025geobench}. Subsequent evaluations broaden the scope beyond plane geometry to 3D settings, larger diagram-based problem spaces, and visually aided mathematical reasoning~\cite{wang2025solidgeo,zhang2026geochallenge,ma2024visaidmath}. Another line pursues formalization and reliable data through verified data construction, formal proof systems, and formal-language-driven synthesis~\cite{fu2025trustgeogen,he2025matpbench,zhang2025geofm}. More recent methods make diagrams less static by incorporating auxiliary construction, geometric transformation, cross-modal rewards, dense sub-goal supervision, and staged reinforcement learning~\cite{weng2025geosketch,guo2025geovlmath,chen2026milestones,wei2025geointr1}. Despite these advances, existing work still mainly operates in symbolic space, where success is defined by recognition, proof, or formal construction. 
Our work bridges symbolic validity and physical execution by grounding geometric reasoning into pixel-space GUI actions that require logical correctness, point-level spatial accuracy, and mitigation of cascading error propagation.

\section{Methodology}
\label{sec:method}

\subsection{Preliminaries}
\label{sec:method_prelim}

We study precision-sensitive geometric GUI drawing, where an agent constructs a target figure on a continuous canvas. Given problem context $Q$ with instruction and target image, the agent starts from canvas state $C_0$ and generates
\begin{equation}
    \tau=(C_0,a_1,C_1,\ldots,a_L,C_L),\quad
    C_\ell=\mathcal{M}(C_{\ell-1},a_\ell),\quad
    a_\ell=(\kappa_\ell,o_\ell,\xi_\ell),
    \label{eq:trajectory}
\end{equation}
where $\mathcal{M}$ is the drawing environment, $\kappa_\ell\in\{\texttt{click},\texttt{paint},\texttt{type}\}$ is the operation type, $o_\ell$ is the object type, and $\xi_\ell$ denotes typed parameters.

The task differs from region-tolerant GUI interaction in success geometry:
\begin{equation}
    \mathrm{Succ}_{\mathrm{reg}}(a)=\mathbb{I}[\mathbf{p}(a)\in R^{*}],
    \qquad
    \mathrm{Succ}_{\mathrm{pt}}(a)=\mathbb{I}[\|\mathbf{p}(a)-\mathbf{p}^{*}\|_2\le\epsilon],
    \label{eq:region_point}
\end{equation}
where $\mathbf{p}(a)$ is the executed pixel location, $R^{*}$ is a valid target region, and $\mathbf{p}^{*}$ is a reference point. Geometric drawing follows the point-level criterion and exhibits dependency-coupled error propagation:
\begin{equation}
    \Delta C_{\ell+1}
    \approx
    \mathbf{J}_{\ell}\Delta C_\ell
    +
    \mathbf{B}_{\ell}\Delta \xi_\ell ,
    \label{eq:error_propagation}
\end{equation}
where $\mathbf{J}_{\ell}$ captures construction dependencies and $\mathbf{B}_{\ell}$ maps parameter errors to canvas perturbations. Thus, small coordinate deviations can affect downstream objects.

\subsection{\textsc{PAGER}: Dependency-Structured Planning and Execution}
\label{sec:method_pager}

\begin{figure*}[t]
    \centering
    \includegraphics[width=\linewidth]{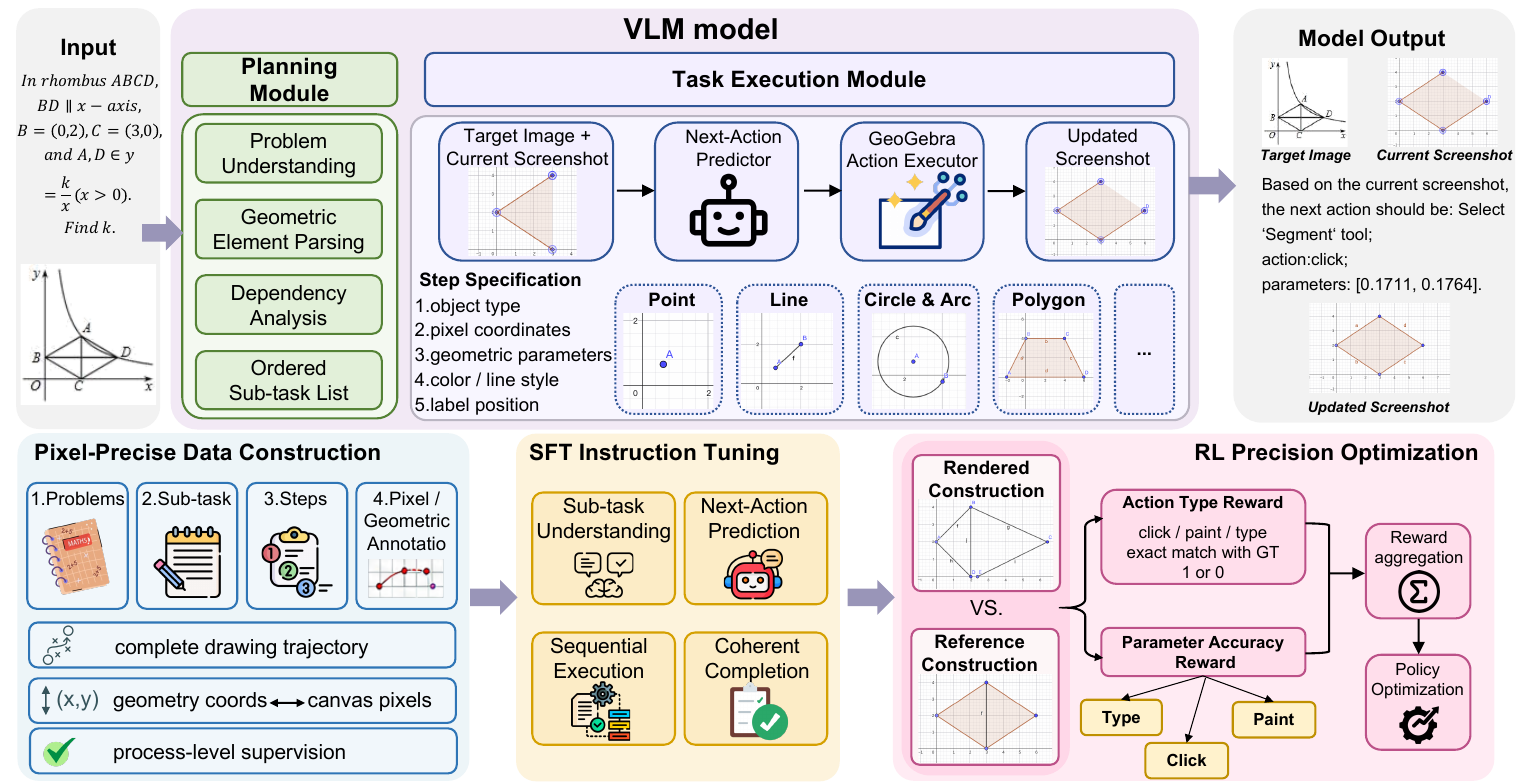}
    \caption{Overview of \textsc{PAGER}. Planning orders sub-tasks, execution grounds them into pixel-level actions, and training aligns supervision with precision rewards.}
    \vspace{-2mm}
    \label{fig:pager_framework}
\end{figure*}

As shown in Figure~\ref{fig:pager_framework}, \textsc{PAGER} factorizes drawing into planning and execution. The Planning Module induces a construction graph and a dependency-consistent sub-task list:
\begin{equation}
    \mathcal{G}_Q=(\mathcal{V}_Q,\mathcal{R}_Q),\quad
    \mathcal{T}=f_\phi(Q)=(T_1,\ldots,T_N),\quad
    (u,v)\in\mathcal{R}_Q^{+}
    \Rightarrow
    \operatorname{rank}_{\mathcal{T}}(u)
    <
    \operatorname{rank}_{\mathcal{T}}(v),
    \label{eq:planning}
\end{equation}
where $\mathcal{V}_Q$ contains primitives or relations, $\mathcal{R}_Q$ encodes dependencies, and $\mathcal{R}_Q^{+}$ is the transitive closure. The Task Execution Module grounds each sub-task into GUI actions:
\begin{equation}
    p_{\phi,\theta}(a_{1:L},\mathcal{T}\mid Q)
    =
    p_{\phi}(\mathcal{T}\mid Q)
    \prod_{i=1}^{N}
    \prod_{j=1}^{m_i}
    \pi_{\theta}
    \big(
    a_{i,j}
    \mid
    Q,T_i,C_{i,j-1},\mathcal{H}_{i,j-1}
    \big),
    \label{eq:hierarchical_policy}
\end{equation}
where $m_i$ is the number of actions for $T_i$, $\mathcal{H}_{i,j-1}$ is action history, nested actions flatten to $a_{1:L}$, and $(o_{i,j},\xi_{i,j})$ instantiates the Step Specification with pixel coordinates, geometric parameters, visual style, and label position.

\subsection{Pixel-Grounded Supervised Tuning}
\label{sec:method_sft}

Pixel-Precise Data Construction provides trajectories $(Q,\mathcal{T}^{*},\tau^{*})$ with sub-tasks, screenshots, histories, next actions, execution feedback, and spatial annotations. For visible window $\Omega=[x_{\min},x_{\max}]\times[y_{\min},y_{\max}]$, geometric coordinates are projected to pixels by:
\begin{equation}
    \Pi_\Omega(x,y)
    =
    \left(
    \frac{x-x_{\min}}{x_{\max}-x_{\min}},
    \frac{y_{\max}-y}{y_{\max}-y_{\min}}
    \right),
    \qquad
    \mathbf{p}
    =
    \operatorname{diag}(W_c,H_c)\Pi_\Omega(x,y),
    \label{eq:geo_pixel_projection}
\end{equation}
where $W_c$ and $H_c$ are canvas width and height. The same projection binds anchors of points, lines, circles, arcs, polygons, and labels to pixel targets. SFT optimizes:
\begin{equation}
    \mathcal{L}_{\mathrm{SFT}}
    =
    -
    \sum_{(Q,\mathcal{T}^{*},\tau^{*})\in\mathcal{D}_{\mathrm{SFT}}}
    \sum_{i=1}^{N^{*}}
    \sum_{j=1}^{m_i^{*}}
    \log
    \pi_{\theta}
    \big(
    a^{*}_{i,j}
    \mid
    Q,T^{*}_i,C^{*}_{i,j-1},\mathcal{H}^{*}_{i,j-1}
    \big),
    \label{eq:sft}
\end{equation}
where $N^{*}=|\mathcal{T}^{*}|$ and $m_i^{*}$ is the number of reference actions for $T_i^{*}$. SFT learns executable action grammar and state-conditioned action prediction, but teacher forcing uses reference screenshots while inference uses self-generated screenshots. Eq.~\ref{eq:error_propagation} therefore motivates precision-aware rollout training.

\subsection{Precision-Aligned Reinforcement Learning}
\label{sec:method_rl}

RL Precision Optimization aligns the policy with action-type correctness, parameter accuracy, and rendered geometric validity. For each problem, the Planning Module produces $\mathcal{T}=f_\phi(Q)$, and policy-environment interaction induces rollout $\hat{\tau}\sim(\pi_\theta,\mathcal{M})(\cdot\mid Q,\mathcal{T})$ with length $\hat{L}$ and rendered construction $\hat{G}=\operatorname{Render}(\hat{\tau})$. Each sampled action is scored against:
\begin{equation}
    \mathcal{A}_{\ell}
    =
    \operatorname{Adm}
    \big(
    Q,\mathcal{T},\hat{C}_{\ell-1},G^{*}
    \big)
    \subseteq
    \{(\kappa,o,\xi):\kappa\in\mathcal{K},\,o\in\mathcal{O},\,\xi\in\Xi_{\kappa,o}\},
    \label{eq:admissible_set}
\end{equation}
where $G^{*}$ is the reference construction, and $\mathcal{K}$, $\mathcal{O}$, and $\Xi_{\kappa,o}$ are operation, object, and typed parameter spaces. The admissible set is built by a training-time geometric verifier and is not used during inference. The rollout reward is:
\begin{equation}
\begin{aligned}
    r_\ell
    &=
    \max_{\tilde{a}\in\mathcal{A}_\ell}
    \mathbb{I}[\hat{\kappa}_\ell=\tilde{\kappa}]
    \left(
    \lambda_a
    +
    \lambda_p
    \exp[-\delta(\hat{a}_\ell,\tilde{a})/\sigma_p]
    \right), \\
    R(\hat{\tau})
    &=
    \frac{1}{\hat{L}}
    \sum_{\ell=1}^{\hat{L}} r_\ell
    +
    \lambda_g
    \exp[-d_{\mathrm{geo}}(\hat{G},G^{*})/\sigma_g].
\end{aligned}
\label{eq:reward}
\end{equation}
For $\tilde{a}=(\tilde{\kappa},\tilde{o},\tilde{\xi})$, operation-type matching grants $\lambda_a$ and activates the parameter-accuracy term. The distance $\delta$ penalizes object mismatch and typed parameter error, including text consistency for \texttt{type}, region validity for \texttt{click}, and pixel deviation for \texttt{paint}; $d_{\mathrm{geo}}$ compares anchors, relations, and layout. The policy is optimized with the SFT policy as a KL anchor:
\begin{equation}
    \max_{\theta}\;
    \mathbb{E}_{\substack{
    Q\sim\mathcal{D}_{\mathrm{RL}},\,\mathcal{T}=f_\phi(Q)\\
    \hat{\tau}\sim(\pi_\theta,\mathcal{M})(\cdot\mid Q,\mathcal{T})
    }}
    \left[
    R(\hat{\tau})
    -
    \beta
    D_{\mathrm{KL}}
    \!\left(
    \pi_{\theta}(\cdot\mid Q,\mathcal{T})
    \,\|\, 
    \pi_{\mathrm{SFT}}(\cdot\mid Q,\mathcal{T})
    \right)
    \right].
    \label{eq:rl_objective}
\end{equation}
The KL term preserves executable behavior, while the reward targets the point-level criterion in Equation~\ref{eq:region_point} and the cascading-error mechanism in Equation~\ref{eq:error_propagation}.

\section{Dataset}

\subsection{Dataset Construction}

\begin{figure}[ht]
    \centering
    \includegraphics[width=\linewidth]{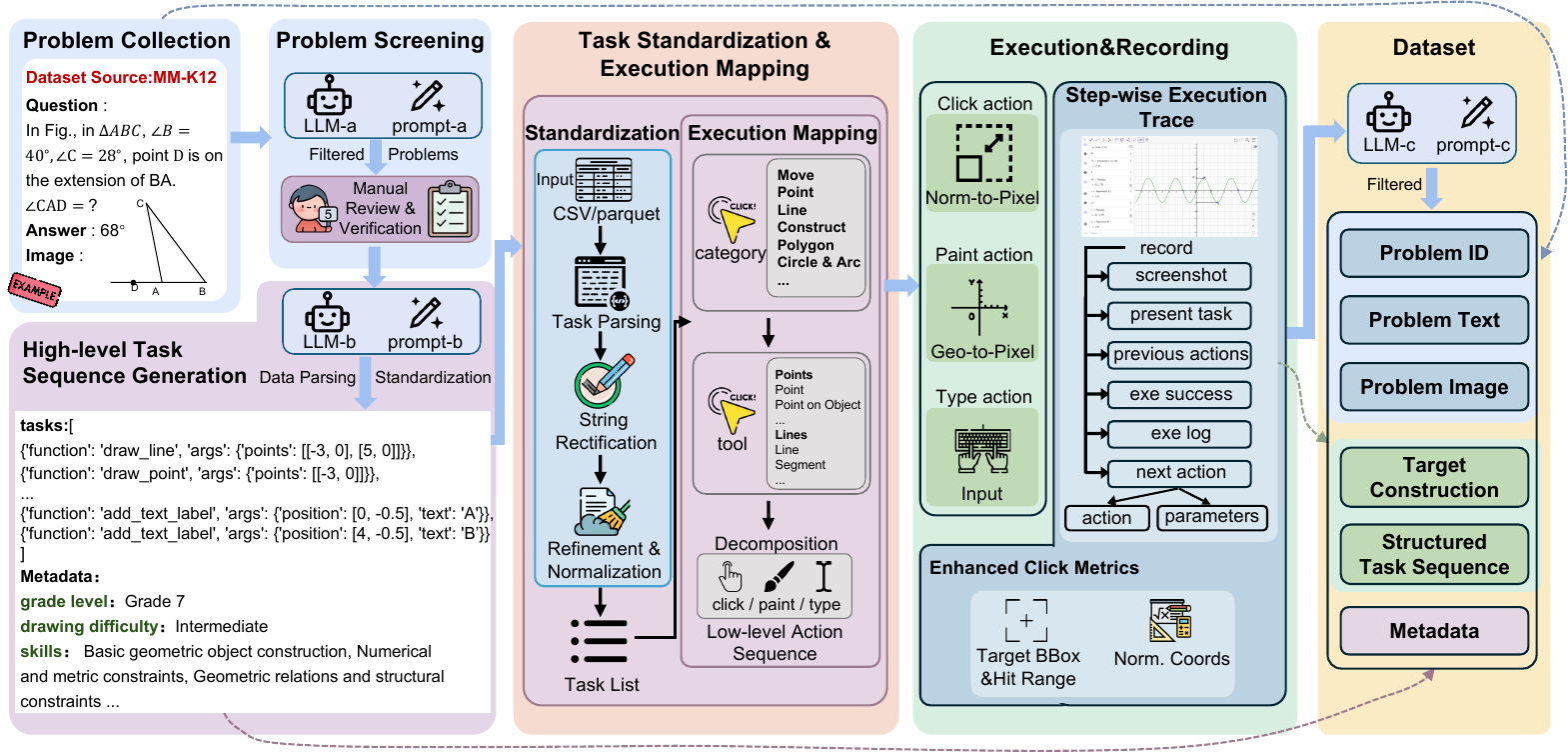}
    \caption{Construction pipeline of \textsc{PAGE} Bench. Candidate geometry problems are screened for GeoGebra-executable instances, converted into structured task sequences, mapped to low-level GUI actions, executed with step-wise recording in a live environment, and finally filtered to retain high-quality trajectories for precision-sensitive geometric GUI learning and evaluation.}
    % \vspace{-2mm}
    \label{fig:dataset_pipeline}
\end{figure}

As shown in Figure~\ref{fig:dataset_pipeline}, \textsc{PAGE} Bench is constructed as a closed execution loop rather than a static collection. The full pipeline converts raw problems into executable construction trajectories in GeoGebra and retains only those instances that remain valid after execution and verification.

\noindent\textbf{Problem collection and executable screening.}
A candidate pool is first assembled from public K--12 multimodal geometry resources~\cite{du2025mm}. Since many raw items support symbolic solving but not GUI-grounded construction, a model-assisted screening module selects problems whose solutions can be realized as ordered constructions in GeoGebra, and manual verification then removes under-specified statements, non-constructive formulations, and cases whose dependencies cannot be operationalized on the canvas. This stage yields construction-ready problems whose solution logic can be grounded in interface actions rather than free-form derivations.

\noindent\textbf{Structured task generation and standardization.}
For each retained problem, a language-model-based authoring module produces a high-level task sequence represented as an ordered list of \texttt{function+args} operations. A subsequent standardization module parses the generated structures, rectifies malformed task strings, and normalizes the output into a canonical task list together with aligned metadata. The result is a structured intermediate representation that makes the intended geometric dependencies and execution order explicit.

\noindent\textbf{Execution mapping and environment-grounded reconstruction.}
The standardized task list is next mapped to low-level GUI interactions in a live GeoGebra environment. Each abstract construction step is decomposed into a sequence of tool-category selection, tool selection, and parameterized canvas manipulation, yielding executable \texttt{click}, \texttt{paint}, and \texttt{type} actions. A unified interaction layer converts structured construction intent into browser-level operations, while coordinate normalization, geometry-to-pixel projection, and boundary-aware retry preserve executability under varying browser states and out-of-canvas conditions. In this way, symbolic construction plans are reconstructed as replayable interface trajectories.

\noindent\textbf{Execution recording, post-execution filtering, and final packaging.}
During execution, the framework records, for each step, the \texttt{screenshot}, \texttt{present task}, \texttt{previous actions}, \texttt{exe success}, \texttt{exe log}, and \texttt{next action}, together with the executed \texttt{action} and its \texttt{parameters}. For \texttt{click} operations, the recorder additionally preserves the target bounding box, hit range, and normalized coordinates, providing the fine-grained spatial evidence required for later precision analysis. After execution, a final language-model-based filtering module compares recorded trajectories against rendered outcomes and removes inconsistent task sequences, failed executions, and geometrically invalid constructions. The retained benchmark therefore provides verified construction trajectories with fine-grained spatial provenance, making it possible to study point-level accuracy and cascading geometric errors in precision-sensitive GUI tasks.

\subsection{Dataset Analysis}

\begin{figure}[ht]
    \centering
    \vspace{-2mm}
    \includegraphics[width=\linewidth]{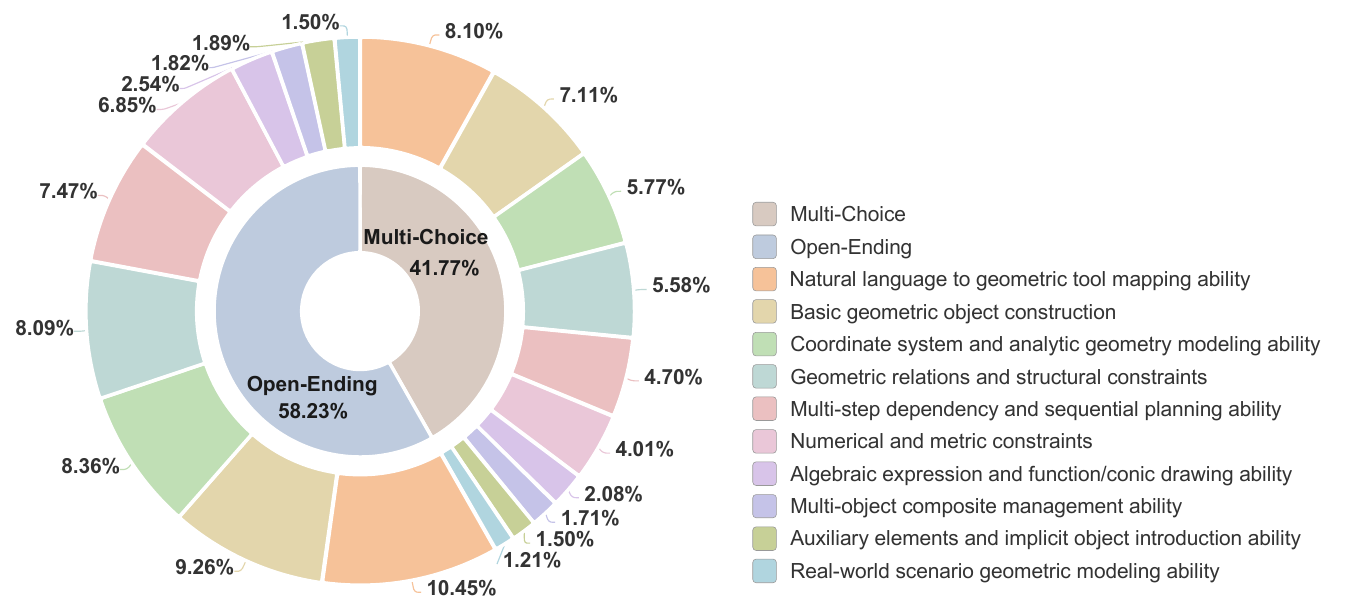}
    \vspace{-1mm}
    \caption{Question-type and skill composition in \textsc{PAGE} Bench.}
    \vspace{-1mm}
    \label{fig:page_category_breakdown}
\end{figure}

Figure~\ref{fig:page_category_breakdown} and Table~\ref{tab:page_dataset_stats} summarize the composition and process scale of \textsc{PAGE} Bench.
\textsc{PAGE} Bench contains 4{,}906 problems with a 4{,}443/463 train-test split, including 2{,}049 multiple-choice and 2{,}857 open-ended instances. The 58.23\% open-ended share emphasizes explicit construction rather than answer selection. Its ten-category multi-label taxonomy yields 25{,}301 annotations, or 5.16 tags per problem, indicating that most instances combine language-to-tool grounding, object construction, coordinate modeling, relation reasoning, multi-step planning, and auxiliary construction rather than isolated skills. Most problems come from Grades 8--10+, and intermediate or hard cases account for 94.11\%, placing the benchmark in a construction-oriented and nontrivial reasoning regime.

\begin{table}[ht]
  \small
  \centering
  \caption{Key statistics of \textsc{PAGE} Bench.}
  \label{tab:page_dataset_stats}
  \renewcommand{\arraystretch}{1.05}
  % \resizebox{\linewidth}{!}{
  \setlength{\tabcolsep}{3.2mm}
  \begin{tabular}{lrrlrr}
    \toprule
    \multicolumn{3}{c}{\textbf{Corpus statistics}} &
    \multicolumn{3}{c}{\textbf{Process statistics}} \\
    \cmidrule(lr){1-3}\cmidrule(lr){4-6}
    Statistic & Count & Share &
    Statistic & Count & Share \\
    \midrule
    Total problems & 4{,}906 & 100.00\% &
    Total tasks & 53{,}277 & -- \\
    Train / Test & 4{,}443 / 463 & 90.56 / 9.44 &
    Avg.\ tasks / problem & 10.86 & -- \\
    Multiple-choice & 2{,}049 & 41.77\% &
    Total actions & 224{,}497 & -- \\
    Open-ended & 2{,}857 & 58.23\% &
    Avg.\ actions / problem & 45.76 & -- \\
    Skill categories & 10 & -- &
    \texttt{click} actions & 107{,}148 & 47.73\% \\
    Category tags & 25{,}301 & 5.16 / prob. &
    \texttt{paint} actions & 90{,}486 & 40.31\% \\
    Grade 8--10+ & 4{,}387 & 89.42\% &
    \texttt{type} actions & 26{,}863 & 11.97\% \\
    Intermediate / Hard & 2{,}940 / 1{,}677 & 94.11\% &
    \texttt{click}+\texttt{paint} & 197{,}634 & 88.03\% \\
    \bottomrule
  \end{tabular}
  % }
\end{table}

On the process side, the corpus contains 53{,}277 high-level tasks and 224{,}497 GUI actions, averaging 10.86 tasks and 45.76 actions per problem. This trajectory length creates meaningful dependency chains, where early execution errors can influence later objects. Spatial operations dominate: \texttt{click} and \texttt{paint} contribute 88.03\% of all actions, with \texttt{paint} directly requiring continuous-canvas control.

\section{Experiment}
\subsection{Experimental Setup}

We train \textsc{PAGER} from Qwen3-VL-8B~\cite{bai2025qwen3}. During supervised fine-tuning, we update the vision encoder, multimodal projector, and language backbone with a maximum input length of 8{,}192 tokens, per-device batch size 1, gradient accumulation 4, learning rate $5\times10^{-6}$, 5\% warmup, bfloat16 precision, and DeepSpeed ZeRO-2 over 8 GPUs for 1 epoch. The reinforcement-learning stage follows SFT and uses rejection sampling with 8 candidates per prompt; prompts with high outcome variance are retained to focus optimization on uncertain rollouts. All stages are implemented with \texttt{torchrun} on 8 NVIDIA A100 GPUs. Evaluation metrics are detailed in Appendix~\ref{metrics}.

We compare against open-source VLMs, including Qwen3-VL-8B~\cite{bai2025qwen3}, DeepSeek-VL2~\cite{wu2024deepseek}, GLM-4.5V~\cite{hong2025glm}, InternVL2.5-8B~\cite{zhu2025internvl3}, KimiVL-A3B~\cite{team2025kimi}, MiniCPM-V-2.6~\cite{yao2024minicpm}, and LLaVA-NeXT-8B~\cite{liu2024llavanext}; closed-source VLMs, including Claude-Sonnet-4.6~\cite{Anthropic2026ClaudeSonnet4_6}, GPT-5.4~\cite{OpenAI2026GPT5_4}, Qwen3.6-Plus~\cite{qwen3.6-35b-a3b}, and Gemini-3.1-Pro~\cite{GoogleDeepMind2026Gemini3_1Pro}; and GUI-specialized agents, including UI-TARS~\cite{qin2025ui}, OS-ATLAS~\cite{wu2024atlas}, InfiGUI-R1-3B~\cite{liu2025infigui}, GUI-Actor-7B~\cite{shakeel2026medspot}, and OpenCUA-7B~\cite{wang2025opencua}. This benchmark set covers general multimodal reasoning, proprietary vision-language modeling, and interface-specialized action prediction.

\subsection{Main Results}

\begin{table*}[ht]

\centering
\footnotesize
\caption{
Main results on precise geometric tasks. Models are grouped into open-source VLMs, closed-source VLMs, specialized GUI agents, and our method. The best are highlighted in bold.
}
\label{tab:main_results}
\setlength{\tabcolsep}{3.5pt}
% \resizebox{\textwidth}{!}{
\begin{tabular}{lccccccccccc}
\toprule
\textbf{Model}

& \textbf{Action} 
& \textbf{Param} 
& \textbf{Step} 
& \textbf{Task} 
& \textbf{O-Comp.} 
& \textbf{S-Comp.} 
& \textbf{S-Vis.} 
& \textbf{S-Geo.} 
& \textbf{Middle} 
& \textbf{Final} 
& \textbf{Overall} \\

\midrule
\multicolumn{12}{c}{\emph{Open-source VLMs}} \\
\midrule
Qwen3-VL-8B~\cite{bai2025qwen3} 
& 52.03 & 9.48 & 9.38 & 0.25 & 3.09 & 3.83 & 3.69 & 3.05 & 8.18 & 3.42 & 5.80 \\
DeepSeek-VL2~\cite{wu2024deepseek} 
& 26.33 & 1.75 & 1.53 & 0.00 & 18.25 & 8.46 & 3.70 & 12.40 & 3.11 & 11.23 & 7.17 \\
GLM-4.5V~\cite{hong2025glm} 
& 78.91 & 12.17 & 11.76 & 0.00 & 5.44 & 7.93 & 2.89 & 13.41 & 11.46 & 7.27 & 9.37 \\
InternVL2.5-8B~\cite{zhu2025internvl3}
& 40.34 & 1.48 & 1.33 & 0.00 & 9.23 & 7.19 & 1.72 & 10.85 & 4.45 & 7.44 & 5.94 \\
KimiVL-A3B~\cite{team2025kimi} 
& 46.60 & 0.59 & 0.57 & 0.00 & 1.70 & 7.12 & 1.24 & 14.05 & 4.83 & 5.70 & 5.27 \\
MiniCPM-V-2.6-8B~\cite{yao2024minicpm}
& 45.87 & 1.06 & 0.71 & 0.00 & 16.01 & 4.81 & 1.17 & 8.18 & 4.84 & 8.12 & 6.48 \\
LLaVA-NeXT-8B~\cite{liu2024llavanext}
& 45.77 & 1.69 & 1.59 & 0.00 & 8.04 & 5.13 & 1.74 & 8.75 & 	5.06 & 6.05 & 5.56 \\

\midrule
\multicolumn{12}{c}{\emph{Closed-source VLMs}} \\
\midrule
Claude-Sonnet-4.6~\cite{Anthropic2026ClaudeSonnet4_6} 
& \cellcolor[rgb]{.851,.953,.992}\textbf{95.85} & 36.44 & 36.38 & 1.11 & 7.04 & 9.46 & 3.11 & 15.39 & 21.17 & 8.65 & 14.91 \\
GPT-5.4~\cite{OpenAI2026GPT5_4} 
& 88.04 & 31.71 & 31.41 & 0.56 & 10.99 & 9.59 & 3.53 & 15.39 & 18.59 & 9.96 & 14.28 \\
Qwen3.6-Plus~\cite{qwen3.6-35b-a3b} 
& \cellcolor[rgb]{.851,.961,.839}90.95 & 51.60 & 51.07 & 4.90 & 18.19 & 11.04 & 4.23 & 10.52 & 27.41 & 11.72 & 19.56 \\
Gemini-3.1-Pro~\cite{GoogleDeepMind2026Gemini3_1Pro} 
& 89.18 & \cellcolor[rgb]{.851,.953,.992}\textbf{66.68} & \cellcolor[rgb]{.851,.953,.992}\textbf{66.66} & \cellcolor[rgb]{.851,.961,.839}5.82 & \cellcolor[rgb]{.851,.961,.839}20.97 & \cellcolor[rgb]{.851,.961,.839}14.17 & \cellcolor[rgb]{.851,.953,.992}\textbf{7.63} & \cellcolor[rgb]{.851,.953,.992}\textbf{21.21} & \cellcolor[rgb]{.851,.961,.839}32.41 & \cellcolor[rgb]{.851,.961,.839}16.31 & \cellcolor[rgb]{.851,.961,.839}24.36 \\

\midrule
\multicolumn{12}{c}{\emph{Specialized GUI Agents}} \\
\midrule
UI-TARS~\cite{qin2025ui} 
& 47.08 & 8.79 & 8.38 & 0.00 & 7.06 & 5.26 & 3.78 & 5.21 & 7.26 & 5.49 & 6.38 \\
OS-ATLAS~\cite{wu2024atlas} 
& 51.24 & 9.19 & 8.80 & 0.29 & 14.56 & 6.65 & 4.32 & 6.36 & 7.98 & 8.50 & 8.24 \\
InfiGUI-R1-3B~\cite{liu2025infigui} 
& 44.81 & 16.51 & 16.18 & 0.00 & 18.23 & 9.66 & 4.14 & 13.82 & 9.37 & 11.96 & 10.66 \\
OpenCUA-7B~\cite{wang2025opencua} 
& 55.86 & 10.57 & 9.85 & 0.12 & 15.39 & 8.92 & 3.11 & 10.77 & 8.69 & 10.07 & 9.38 \\
GUI-Actor-7B~\cite{shakeel2026medspot} 
& 47.26 & 5.31 & 5.02 & 0.00 & 18.66 & 6.04 & 1.42 & 7.58 & 6.26 & 9.21 & 7.74 \\

\midrule
% \multicolumn{12}{l}{\emph{Ours}} \\
\textbf{PAGER} 
& 82.62 & \cellcolor[rgb]{.851,.961,.839}62.76 & \cellcolor[rgb]{.851,.961,.839}62.20 & \cellcolor[rgb]{.851,.953,.992}\textbf{23.78} & \cellcolor[rgb]{.851,.953,.992}\textbf{28.88} & \cellcolor[rgb]{.851,.953,.992}\textbf{15.30} & \cellcolor[rgb]{.851,.961,.839}7.05 & \cellcolor[rgb]{.851,.961,.839}15.63 & \cellcolor[rgb]{.851,.953,.992}\textbf{41.25} & \cellcolor[rgb]{.851,.953,.992}\textbf{17.79} & \cellcolor[rgb]{.851,.953,.992}\textbf{29.52} \\
\bottomrule

\end{tabular}
% }
\vspace{-4mm}
\end{table*}

Table~\ref{tab:main_results} shows that \textsc{PAGER} achieves the best Overall score, 29.52, improving over the strongest general baseline, Gemini-3.1-Pro, by 5.15 points, or 21.1\%. It also obtains the highest Task, Middle, and Final scores, indicating stronger complete-rollout execution and better final geometric quality. Notably, these gains occur despite Gemini-3.1-Pro leading in Param and Step, suggesting that \textsc{PAGER} better converts local execution into task-level success. \textbf{This indicates stronger trajectory-level stability rather than merely better single-step prediction.}

The results expose a clear Semantic-Execution Gap. Closed-source VLMs often select the correct operation type: Claude-Sonnet-4.6 reaches 95.85 Action Accuracy, while GPT-5.4 and Gemini-3.1-Pro reach 88.04 and 89.18. However, their Task Success remains 1.11, 0.56, and 5.82, respectively. In contrast, \textsc{PAGER} reaches 23.78 Task Success, about $4.1\times$ Gemini-3.1-Pro. \textbf{This shows that precise drawing is not bottlenecked by action semantics alone, but by state-conditioned parameter control and error accumulation across dependent construction steps.}

Compared with GUI-specialized agents, \textsc{PAGER} further highlights the limitation of region-tolerant GUI. UI-TARS and OS-ATLAS remain below 9\% Step Success, and the strongest GUI-agent baseline reaches only 16.18, whereas \textsc{PAGER} reaches 62.20. This indicates that component-level GUI grounding is too coarse for geometric construction, where exact points, rather than regions, determine validity. These results support the central motivation of this work: \textbf{precision-sensitive geometric GUI control requires point-level execution, geometry-aware feedback, and robustness to cascading coordinate errors, rather than only component-level grounding.}

\subsection{Ablation Study}

Table~\ref{tab:ablation} analyzes the effect of the proposed components. \textsc{PAGER}-SFT already provides a strong execution prior, reaching 48.47 Parameter Accuracy, 47.91 Step Success, and 20.47 Overall, which confirms the value of pixel-grounded process supervision for learning executable drawing grammar. The reward ablations show that parameter alignment is the key bottleneck. Without the parameter-accuracy reward, the model gains little beyond SFT and even drops from 20.47 to 20.07 Overall, suggesting that action-level correctness alone cannot preserve geometric structure when coordinates, endpoints, radii, or labels drift. Without the action-type reward, the model still improves to 24.52 Overall and 15.90 Task Success, indicating that continuous-space precision is central to valid construction. The full \textsc{PAGER} combines both rewards and achieves the best performance across most metrics, improving Overall from 20.47 to 29.52 and Task Success from 4.48 to 23.78 over SFT. The two rewards are therefore complementary: action-type reward stabilizes semantic execution order, while parameter-accuracy reward improves point-level control. Their combination yields the most reliable alignment between action selection, parameter prediction, and final geometric validity.

\begin{table*}[ht]
\centering
\vspace{-3mm}
\caption{
Ablation study of different training strategies. The best results are highlighted in bold.
}
\footnotesize
\renewcommand{\arraystretch}{1.25}
\label{tab:ablation}
\setlength{\tabcolsep}{4.0pt}
% \resizebox{\textwidth}{!}{
\begin{tabular}{lccccccccccc}
\toprule
\textbf{Model} 
& \textbf{Action}
& \textbf{Param}
& \textbf{Step}
& \textbf{Task}
& \textbf{O-Comp.}
& \textbf{S-Comp.}
& \textbf{S-Vis.}
& \textbf{S-Geo.}
& \textbf{Middle}
& \textbf{Final}
& \textbf{Overall} \\
\midrule
PAGER-SFT
& 75.14 & 48.47 & 47.91 & 4.48
& 27.70 & 11.24 & 5.69 & \cellcolor[rgb]{.851,.953,.992}\textbf{17.48}
& 24.63 & 16.32 & 20.47 \\

PAGER o/w RL$_{\text{param}}$
& 74.72 & 52.88 & 52.40 & 5.62
& 20.11 & 10.99 & 4.35 & 16.64
& 26.61 & 13.53 & 20.07 \\

PAGER o/w RL$_{\text{action}}$
& 78.77 & 59.14 & 58.70 & 15.90
& 22.67 & 11.54 & 4.33 & 14.22
& 35.07 & 13.97 & 24.52 \\
\hline
\textbf{PAGER}
& \cellcolor[rgb]{.851,.953,.992}\textbf{82.62} & \cellcolor[rgb]{.851,.953,.992}\textbf{62.76} & \cellcolor[rgb]{.851,.953,.992}\textbf{62.20} & \cellcolor[rgb]{.851,.953,.992}\textbf{23.78}
& \cellcolor[rgb]{.851,.953,.992}\textbf{28.88} & \cellcolor[rgb]{.851,.953,.992}\textbf{15.30} & \cellcolor[rgb]{.851,.953,.992}\textbf{7.05} & 15.63
& \cellcolor[rgb]{.851,.953,.992} \cellcolor[rgb]{.851,.953,.992}\textbf{41.25} & \cellcolor[rgb]{.851,.953,.992}\textbf{17.79} & \cellcolor[rgb]{.851,.953,.992}\textbf{29.52} \\
\bottomrule
\end{tabular}
\vspace{-3mm}
% }
\end{table*}

\subsection{Case Study and Error Analysis}
\label{sec:case_study}

\begin{figure}[t]
    \vspace{-5mm}
    \centering
    \includegraphics[width=1\linewidth]{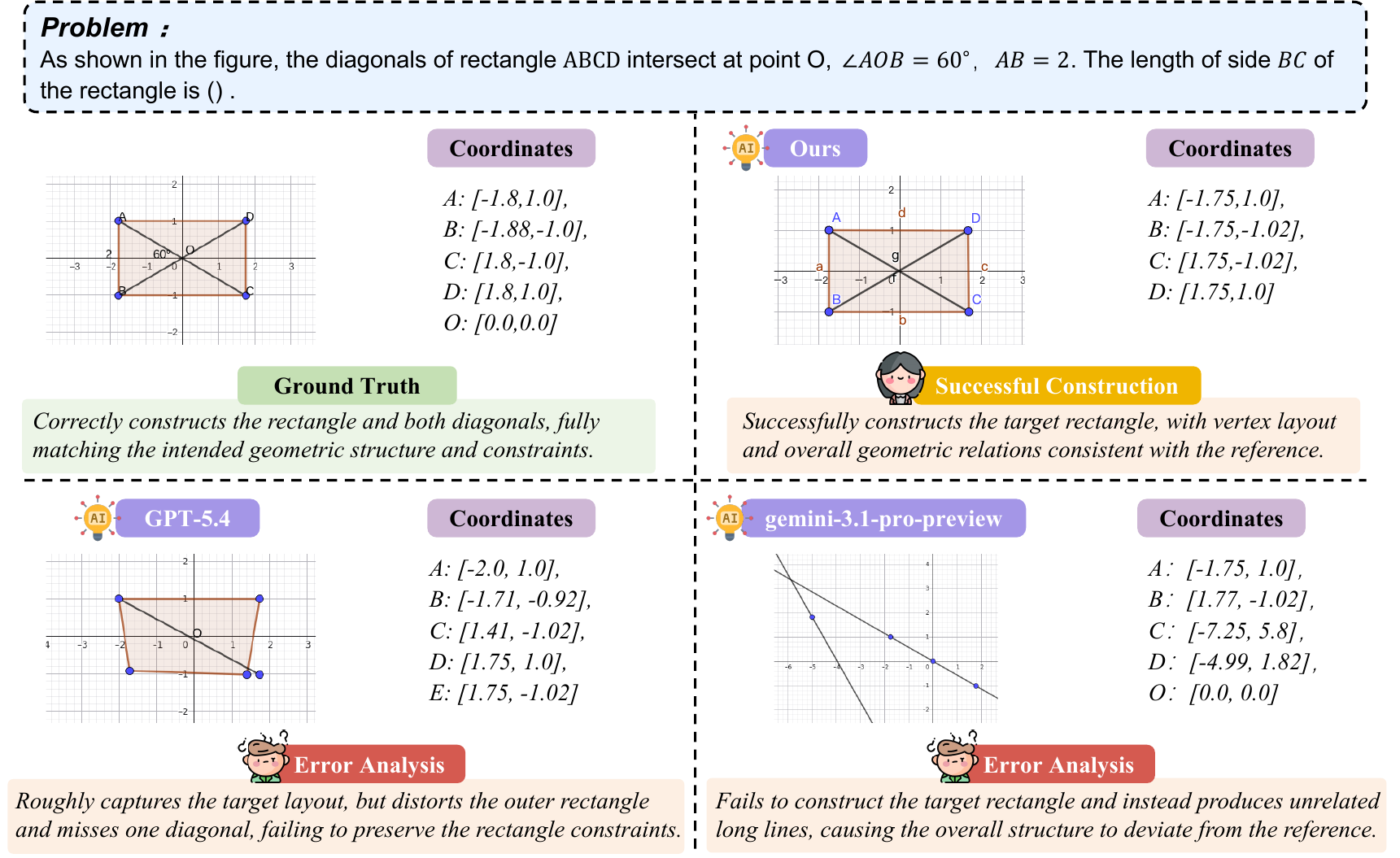}
    \vspace{-5mm}
    \caption{Qualitative comparison. \textsc{PAGER} better preserves rectangular structure, diagonal intersection, and coordinate consistency.}
    \vspace{-5mm}
    \label{fig:case_study}
\end{figure}

Figure~\ref{fig:case_study} presents a representative construction task: drawing rectangle $ABCD$ with diagonals intersecting at $O$, under $\angle AOB=60^\circ$ and $AB=2$, and determining $BC$. The example requires accurate vertex placement, diagonal construction, and preservation of side, angle, and intersection constraints. \textsc{PAGER} constructs a geometrically consistent rectangle with valid side relations and central diagonal intersection, despite minor numerical deviations. GPT-5.4 captures the rough intent but distorts the quadrilateral, introduces redundant elements, and omits a key diagonal. Gemini-3.1-Pro exhibits stronger parameter drift: early vertex errors propagate into invalid long segments and break the rectangular structure. This case illustrates that failures in precise drawing often arise from unstable pixel-level parameters and weak constraint preservation, rather than from complete semantic misunderstanding. It therefore supports the quantitative finding that \textsc{PAGER}'s process supervision and precision-aligned optimization improve execution reliability under strict geometric constraints.

\subsection{Consistency with Human Judgments}
\label{sec:consistency_human}

\noindent
\begin{minipage}{0.62\columnwidth}
    Figure~\ref{fig:human_consistency} illustrates a clear spatial separation in model performance. Most existing MLLMs, including GPT-5.4 and Gemini-3.1-Pro, cluster in the lower-left region with both low automated scores and low human ratings.  In contrast, \textsc{PAGER} occupies the top-right corner, achieving high automated success alongside superior human preference. The near-perfect correlation ($r=0.9397$) demonstrates a strong alignment between our automated verification and expert judgment. Despite the precision-sensitive nature of the task, improvements in our metrics translate monotonically into human-perceived correctness, confirming that \textsc{PAGE} Bench captures genuine geometric validity rather than proxy signals.
\end{minipage}
\hfill
\begin{minipage}{0.35\columnwidth}
    \centering
    \includegraphics[width=\linewidth]{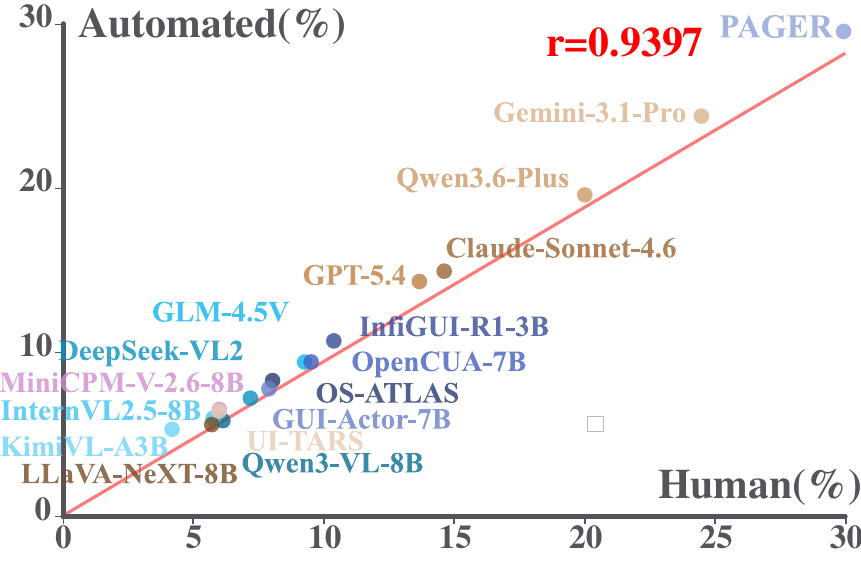}
    \captionof{figure}{Automatic evaluation score vs.\ human rating.}
    \label{fig:human_consistency}
\end{minipage}

\section{Conclusion and Limitations}

We introduce \emph{precision-sensitive GUI tasks}, where GUI success shifts from region-tolerant component selection to point-precise control in continuous visual space. Precise geometric construction exposes this regime because dependency-coupled primitives allow small coordinate errors to cascade into invalid downstream structures. \textsc{PAGE} Bench makes this failure mode measurable with process-supervised, pixel-level geometric interactions, and \textsc{PAGER} addresses it through dependency-structured planning and precision-aligned execution. Experiments reveal a clear \textit{Semantic-Execution Gap}: existing agents often identify the correct operation but fail to realize it with the spatial precision required by geometry, while \textsc{PAGER} narrows this gap and advances GUI agents toward faithful execution of structured intent in pixel space. This work focuses on GeoGebra-style planar construction, so other precision-sensitive interfaces may require additional action grammars, environment adapters, and validity rules. This controlled setting isolates the core semantic-execution gap and provides verifier-backed supervision, while extending the same principle to broader domains such as CAD, diagram editing, and scientific visualization remains a natural direction.

% \bibliography{ref}
% \bibliographystyle{unsrt}

% \clearpage
\newpage
\bibliographystyle{plainnat}
\setcitestyle{numbers}
\bibliography{ref}

\clearpage
\newpage
\appendix
\section{Performance Results with Radar Chart Analysis}
\label{app:classification_results}

\begin{figure}[ht]
    \centering
    \includegraphics[width=0.8\linewidth]{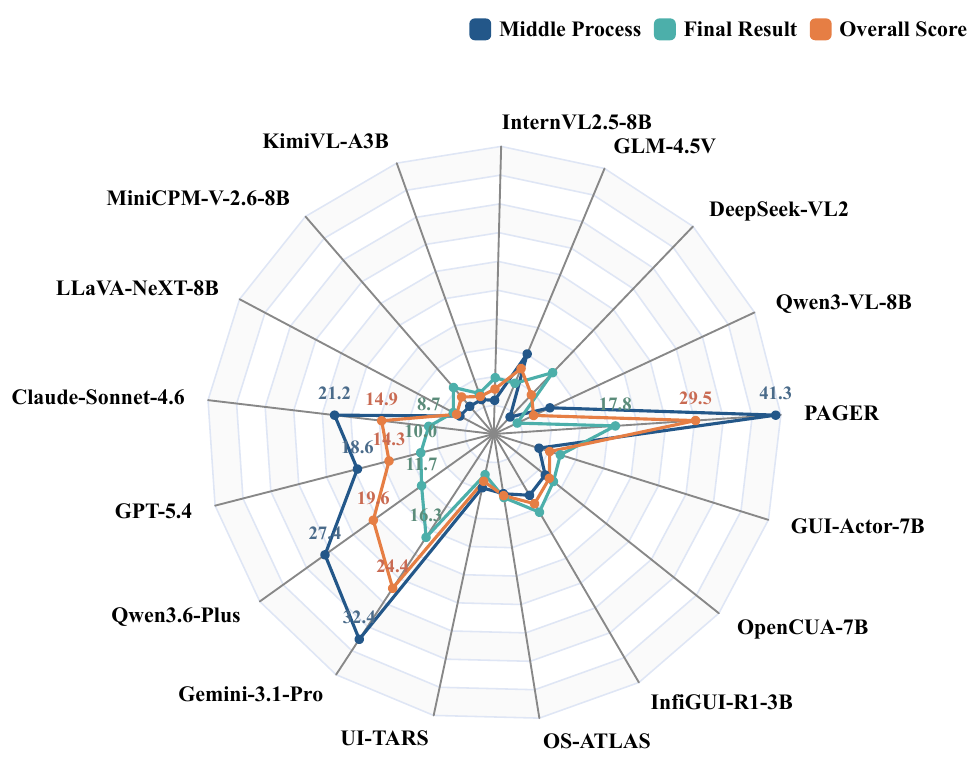}
    \caption{Performance comparison of \textsc{PAGER} against fourteen baselines on \textsc{PAGE} Bench.}
    \vspace{-3mm}
    \label{fig:radar_chart}
\end{figure}

To analyze model performance on precision-sensitive GUI tasks, we evaluate \textsc{PAGER} against fourteen baselines across three categories, as shown in Figure~\ref{fig:radar_chart}. The evaluation reveals a critical Semantic-Execution Gap in existing architectures. State-of-the-art closed models like Gemini-3.1-Pro and GPT-5.4 show strong high-level semantic planning, achieving Middle Process scores of 32.4 and 27.4 respectively. However, their Final Result scores drop drastically, with GPT-5.4 falling to 11.8. This steep degradation highlights that advanced general models lack the fine-grained continuous spatial reasoning needed to prevent dependency-driven error propagation. Leading open-weight architectures, including Qwen3-VL-8B and InternVL2.5-8B, cluster near the radar chart center, reflecting systemic struggles across all metrics. Similarly, GUI-specialized agents designed for interface automation, such as UI-TARS and OS-ATLAS, exhibit severely limited performance. Trained primarily on region-tolerant paradigms, these agents fail to adapt to tasks requiring strict pixel-level parameter accuracy and geometry-aware verification. Our proposed framework consistently outperforms all evaluated baselines. By factorizing construction into dependency-structured planning and pixel-level execution, \textsc{PAGER} achieves the highest Middle Process score of 41.3. Furthermore, it effectively closes the Semantic-Execution Gap, delivering a leading Final Result score of 17.8 and an Overall Score of 29.5. These results empirically validate that our precision-aligned reinforcement learning pipeline successfully maintains spatial accuracy and resists cascading topological failures.

\section{Fine-Grained Classification Results Analysis}
\label{app:classification_analysis}

\begin{figure}[ht]
    \centering
    \includegraphics[width=1\linewidth]{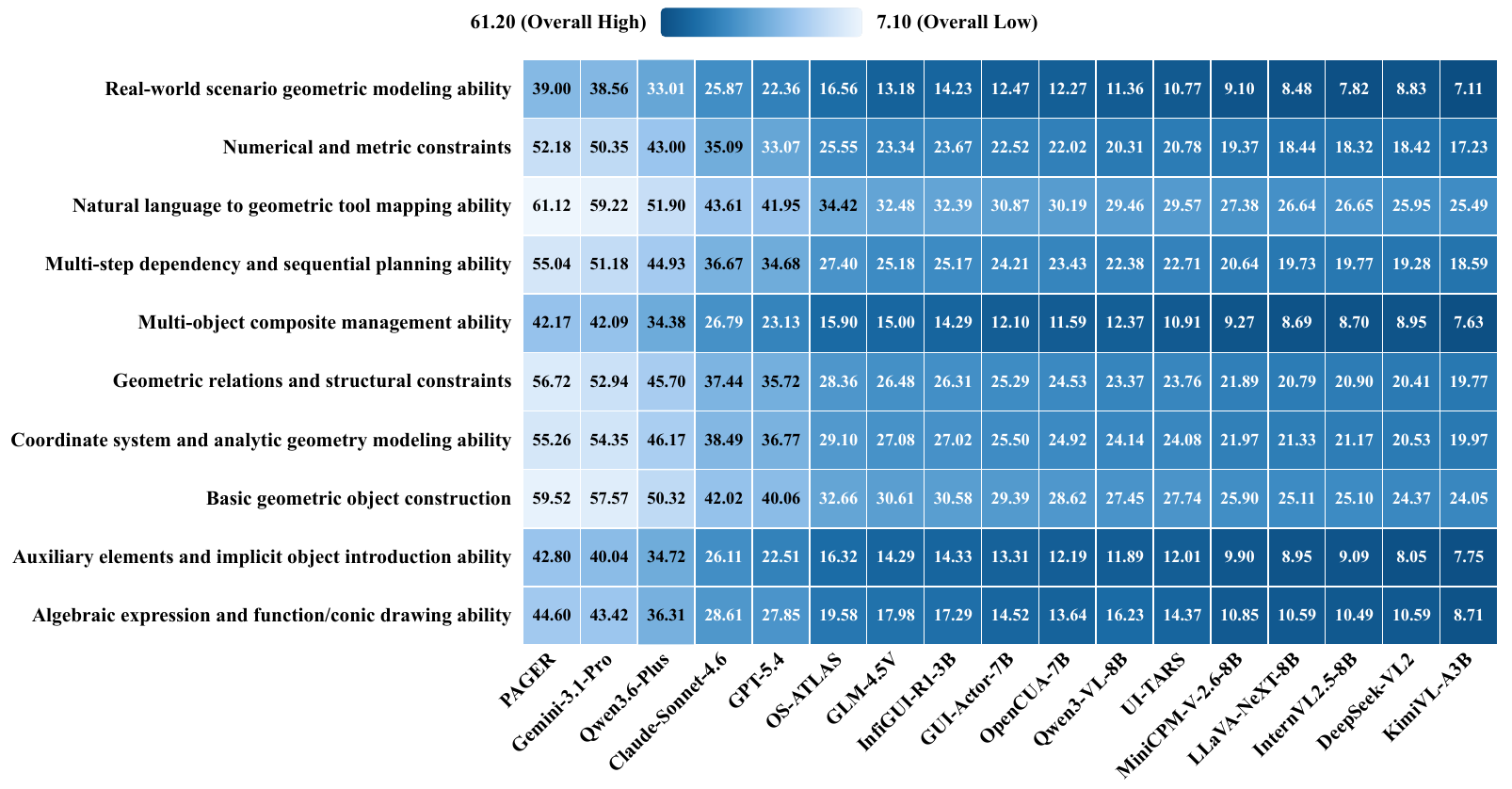}
    \caption{Fine-grained performance breakdown across ten geometric capabilities.}
    \vspace{-3mm}
    \label{fig:heatmap}
\end{figure}

To further investigate the specific bottlenecks of existing models on precision-sensitive GUI tasks, we decompose the evaluation into ten distinct geometric capabilities. Figure~\ref{fig:heatmap} illustrates the performance distribution across these dimensions, highlighting severe variances in how different architectures handle continuous spatial reasoning and ontological dependencies.

Capabilities including natural language to geometric tool mapping and basic geometric object construction represent the fundamental prerequisites for successful drawing. Our proposed \textsc{PAGER} achieves peak performance in these areas with scores of 61.12 and 59.52 respectively. Strong proprietary models like Gemini-3.1-Pro and Qwen3.6-Plus also show reasonable competence here. This indicates that grounding basic semantic instructions to specific drawing tools remains relatively manageable for advanced vision-language models, aligning with their established strengths in semantic parsing.

As evaluation shifts toward multi-step dependency and sequential planning ability alongside geometric relations and structural constraints, baseline performance drops precipitously. GPT-5.4 scores only 34.68 on sequential planning, while open-weight models like Qwen3-VL-8B drop below 23. This degradation empirically validates our hypothesis regarding dependency-coupled operations. Small coordinate deviations in early steps accumulate into cascading topological failures that standard region-tolerant agents cannot resolve. By explicitly factorizing construction into dependency-structured planning and pixel-level execution, \textsc{PAGER} successfully preserves these structural constraints and maintains a sequential planning score of 55.04.

The most pronounced capability gaps emerge in complex scenarios, specifically auxiliary elements and implicit object introduction, multi-object composite management, and real-world scenario geometric modeling. These categories require abstract spatial reasoning and strict geometry-aware verification. GUI-specialized agents like UI-TARS and OS-ATLAS experience near-complete failures in these domains, with scores consistently falling to near 10. Conversely, \textsc{PAGER} demonstrates robust adaptability in these challenging regimes. The precision-aligned reinforcement learning phase effectively mitigates rollout-induced exposure bias, enabling our agent to handle implicit auxiliary constructions and complex multi-object dependencies far more reliably than existing general multimodal paradigms.

\section{GeoGebra Screening Prompt}

\begin{promptbox}[K12-GeoGebra Screening Prompt]
\sloppy\setlength{\emergencystretch}{1em}

\textbf{1. Core Task}

Screen K12 mathematics questions to identify \textbf{geometry-related ones that can use GeoGebra software for graphing to assist problem-solving/teaching}. Exclude questions that do not require or cannot be visualized with GeoGebra, such as pure algebraic calculations or pure logical reasoning.

\vspace{2pt}
\textbf{2. Input Requirements}

\begin{enumerate}[leftmargin=*, itemsep=1pt]
  \item \textbf{Input Format}: Batch text of K12 mathematics questions, which may include stems, options, and problem-solving steps. Separate questions with blank lines or serial numbers.
  \item \textbf{Input Scope}: K12 mathematics questions, including primary, middle, and high school. Specify the school stage if available; otherwise, identify it automatically.
\end{enumerate}

\vspace{2pt}
\textbf{3. Screening Criteria (Applicable if Any Is Met)}

\begin{enumerate}[leftmargin=*, itemsep=1pt]
  \item Core involves \textbf{geometry graph drawing/analysis}: triangles, quadrilaterals, circles, polygons, 3D shapes, coordinate system graphs, etc.
  \item Requires \textbf{graph measurement/construction}: side lengths, angles, areas, volumes, perpendicular/parallel lines, angle bisectors, circumscribed/inscribed circles, and transformations.
  \item Involves \textbf{geometry relationship verification}: congruence, similarity, parallelism, perpendicularity, collinearity, or concyclicity requiring visualization.
  \item Coordinate-system related: point coordinates, linear/circle equations, conic sections, and graph-based property analysis.
\end{enumerate}

\vspace{2pt}
\textbf{4. Exclusion Criteria (Not Applicable if Any Is Met)}

\begin{enumerate}[leftmargin=*, itemsep=1pt]
  \item Pure algebraic calculations: solving equations, factorization, formula calculation, sequences, or non-geometric probability/statistics.
  \item Pure logical reasoning: text-only geometry theorem proofs or definition discrimination.
  \item No clear geometric elements: numerical calculations only or application questions with no graph correlation.
\end{enumerate}

\vspace{2pt}
\textbf{5. Output Format (JSON Structure)}

Output a \textbf{JSON array}, where each element represents a screened question. Each element contains the following fields:

\begin{itemize}[leftmargin=*, itemsep=1pt]
  \item \texttt{school\_stage}: String. School stage of the question: ``Primary'', ``Middle'', or ``High''.
  \item \texttt{can\_draw}: Boolean. \texttt{true} means applicable to GeoGebra; \texttt{false} means not applicable.
  \item \texttt{ggb\_content}: String. If \texttt{can\_draw=true}, provide detailed picture content, including elements and purpose. If \texttt{can\_draw=false}, return ``N/A''.
  \item \texttt{supplementary\_note}: String. Optional. For drawable questions, provide special graph tips. For non-drawable questions, provide the exclusion reason.
\end{itemize}

\vspace{2pt}
\textbf{6. JSON Output Example}

{\footnotesize
\begin{lstlisting}
[
  {
    "school_stage": "Middle",
    "can_draw": true,
    "ggb_content": "Right triangle ABC with right angle at C.
    Mark side AB=5cm. Label vertices A, B, C and right angle
    symbol. And then calculate BC.",
    "supplementary_note": "Use GeoGebra's Right Triangle tool
    for quick construction."
  },
  {
    "school_stage": "Primary",
    "can_draw": false,
    "ggb_content": "N/A",
    "supplementary_note": "Excluded: pure algebraic calculation."
  }
]
\end{lstlisting}
}

\end{promptbox}

\captionof{figure}{Prompt used to screen K12 mathematics questions for GeoGebra-based geometric visualization. The model determines whether each question can be drawn or assisted through GeoGebra and outputs a structured JSON result.}
\label{fig:geogebra_screening_prompt}

\section{GeoGebra Automation Prompt}

\begin{promptbox}[GeoGebra Automation Architect and Dataset Generator]
\sloppy\setlength{\emergencystretch}{1em}

You are a \textbf{GeoGebra Automation Architect and Dataset Generator}. Your mission is to convert high-level mathematical construction descriptions from problem text, answer hints, and image context into a sequence of specific Python function calls based on a strictly defined API library, while categorizing mathematical skills, grade level, and drawing difficulty.

\vspace{2pt}
\textbf{1. Mission Statement}

You will receive a dataset entry containing three core components: \textbf{Question}, \textbf{Answer}, and \textbf{Image}. Your primary focus is to integrate all three components to infer accurate construction steps.

You must:

\begin{enumerate}[leftmargin=*, itemsep=1pt]
  \item \textbf{Analyze}: Analyze the construction logic by combining the problem requirements, answer clues, and image context.
  \item \textbf{Infer}: If coordinates are not explicitly provided, infer reasonable Cartesian coordinates that satisfy the described geometric relationships.
  \item \textbf{Map}: Map each construction step to the exact function in the Allowed Function Library. Custom functions are not permitted.
  \item \textbf{Classify}: Classify the problem by selecting skills, grade level, and drawing difficulty.
  \item \textbf{Labeling rule}: Do not add labels beyond what is required by the Question, Answer, or Image.
  \item \textbf{Output}: Strictly follow the Mandatory Output Format as a JSON object.
\end{enumerate}

\textbf{Note: Replicate the original figure as the highest priority.}

\begin{itemize}[leftmargin=*, itemsep=1pt]
  \item \textbf{Orientation and structure}: The final drawing must match the original image's orientation and structure as closely as possible. Do not rotate, mirror, flip, or re-order the figure.
  \item \textbf{Identifiers}: Replicate all identifiers exactly as shown in the Image, Question, or Answer, including point names, line or segment names, numbering, and explicit marks.
  \item \textbf{Coordinate-plane annotations}: Do not add coordinate labels, axis letters, tick marks, or scale numbers unless they explicitly appear in the Image, Question, or Answer.
  \item \textbf{Auxiliary lines}: Auxiliary lines or segments are allowed if needed to match the original structure or generate required intersection points.
\end{itemize}

\vspace{2pt}
\textbf{2. Allowed Function Library}

You may only use the functions defined below.

\textbf{A. General and Input}
\begin{itemize}[leftmargin=*, itemsep=1pt]
  \item \texttt{generate\_input\_action}: Used for algebraic text input. Parameters: \texttt{\{"text": "string"\}}.
  \item \texttt{add\_text\_label}: Add a text label at a specified position. Parameters: \texttt{\{"position": [x, y], "text": "string"\}}.
\end{itemize}

\textbf{B. Points}
\begin{itemize}[leftmargin=*, itemsep=1pt]
  \item \texttt{draw\_point}: Draw one or more points. Parameters: \texttt{\{"points": [[x1, y1], [x2, y2], ...]\}}.
  \item \texttt{midpoint\_or\_center}: Construct a midpoint or circle center. Parameters: \texttt{\{"points": [[x1, y1], [x2, y2]]\}}.
\end{itemize}

\textbf{C. Lines}
\begin{itemize}[leftmargin=*, itemsep=1pt]
  \item \texttt{draw\_segment}: Draw a finite line segment. Parameters: \texttt{\{"points": [[x1, y1], [x2, y2]]\}}.
  \item \texttt{draw\_line}: Draw an infinite line. Parameters: \texttt{\{"points": [[x1, y1], [x2, y2]]\}}.
  \item \texttt{draw\_ray}: Draw a ray. Parameters: \texttt{\{"points": [[x1, y1], [x2, y2]]\}}.
\end{itemize}

\textbf{D. Construction}
\begin{itemize}[leftmargin=*, itemsep=1pt]
  \item \texttt{perpendicular\_line}: A line through a point and perpendicular to a given line or segment.
  \item \texttt{parallel\_line}: A line through a point and parallel to a given line or segment.
  \item \texttt{perpendicular\_bisector}: Perpendicular bisector of a segment or two points.
  \item \texttt{angle\_bisector}: Angle bisector defined by three points or two lines.
  \item \texttt{tangents}: Tangents to a curve.
\end{itemize}

\textbf{E. Polygons and Shapes}
\begin{itemize}[leftmargin=*, itemsep=1pt]
  \item \texttt{draw\_polygon}: Draw a closed polygon.
  \item \texttt{draw\_circle\_center\_point}: Draw a circle defined by center and a point on the circle.
  \item \texttt{semicircle}: Draw a semicircle defined by diameter endpoints.
  \item \texttt{circular\_sector}: Draw a sector defined by center, start point, and end point.
\end{itemize}

\textbf{F. Conics}
\begin{itemize}[leftmargin=*, itemsep=1pt]
  \item \texttt{parabola}: Defined by a focus and a point on the directrix.
  \item \texttt{hyperbola}: Defined by focus 1, focus 2, and a point on the curve.
\end{itemize}

\vspace{2pt}
\textbf{3. Object Grounding and Dependency Rules}

\begin{enumerate}[leftmargin=*, itemsep=1pt]
  \item \textbf{No floating reference points}: Any reference point used in a construction function must have been previously created or must be strictly located on a previously created object.
  \item \textbf{Create before use}: If a step requires construction based on an existing object, that object must have been created in an earlier task.
  \item \textbf{Strict on-object point constraints}: Coordinates of an on-object point must exactly satisfy the geometric equation of that object.
  \item \textbf{Prefer reuse of existing key points}: Reuse endpoints, vertices, or centers whenever possible.
  \item \textbf{Implicit object registry}: Maintain an internal list of created objects and ensure all later steps reference them consistently.
\end{enumerate}

\textbf{Construction Function Requirements}
\begin{itemize}[leftmargin=*, itemsep=1pt]
  \item \texttt{perpendicular\_line} and \texttt{parallel\_line}: the first point must lie on an existing line, segment, or ray; the second point must be an existing point.
  \item \texttt{perpendicular\_bisector}: both points must be existing points.
  \item \texttt{angle\_bisector}: all three points must already exist, and the middle point is the vertex of the angle.
  \item \texttt{tangents}: input coordinates must reference existing objects and never be free-floating.
\end{itemize}

\vspace{2pt}
\textbf{4. Coordinate Inference Protocol}

\begin{itemize}[leftmargin=*, itemsep=1pt]
  \item If coordinates are not provided, infer reasonable Cartesian coordinates that preserve geometric properties.
  \item \textbf{Integer Priority Rule}: If special coordinates are not required, choose integer coordinates whenever possible.
  \item Use decimals only when integers cannot satisfy the required geometric constraints.
  \item If the figure implies a specific structure, such as a horizontal base or centered vertex, align coordinates with the image context.
\end{itemize}

\vspace{2pt}
\textbf{5. Classification Standards}

\textbf{A. Skill Taxonomy}

Multiple selections are allowed.

\begin{enumerate}[leftmargin=*, itemsep=1pt]
  \item Basic geometric object construction
  \item Numerical and metric constraints
  \item Geometric relations and structural constraints
  \item Multi-step dependency and sequential planning ability
  \item Auxiliary elements and implicit object introduction ability
  \item Coordinate system and analytic geometry modeling ability
  \item Algebraic expression and function/conic drawing ability
  \item Multi-object composite management ability
  \item Natural language to geometric tool mapping ability
  \item Real-world scenario geometric modeling ability
\end{enumerate}

\textbf{B. Grade Level}

Choose one from the following:

\begin{itemize}[leftmargin=*, itemsep=1pt]
  \item Grade 6: basic geometry, points, lines, simple polygons
  \item Grade 7: triangles, quadrilaterals, basic constructions
  \item Grade 8: perpendicular/parallel lines, angle bisectors, circles
  \item Grade 9: advanced constructions, basic conics, polygon properties
  \item Grade 10+: complex conics, theorem application, 2D constructions related to 3D
\end{itemize}

\textbf{C. Drawing Difficulty}

Choose one from the following:

\begin{itemize}[leftmargin=*, itemsep=1pt]
  \item \textbf{Beginner}: 1--2 simple operations.
  \item \textbf{Intermediate}: 3--4 operations involving constructions.
  \item \textbf{Advanced}: 5+ operations or complex logic.
\end{itemize}

\vspace{2pt}
\textbf{6. Mandatory Output Format}

\begin{lstlisting}
[
  {
    "description": "string (brief summary of the construction)",
    "grade_level": "string (chosen from grade levels)",
    "drawing_difficulty": "string (beginner/intermediate/advanced)",
    "skills": ["Skill 1", "Skill 2", "..."],
    "tasks": [
      {
        "function": "FUNCTION_NAME (from allowed function library)",
        "args": {
          "arg_name": "value matching function parameters"
        }
      }
    ]
  }
]
\end{lstlisting}

\vspace{2pt}
\textbf{7. Few-Shot Examples}

\textbf{Example Input 1}
\begin{itemize}[leftmargin=*, itemsep=1pt]
  \item \textbf{Question}: Draw a square with side length 2.
  \item \textbf{Answer}: A square has four right angles and four equal sides.
  \item \textbf{Image}: Suggests a square with a horizontal base.
\end{itemize}

\textbf{Example Output 1}
\begin{lstlisting}
[
  {
    "description": "Construct a square with side length 2",
    "grade_level": "Grade 7",
    "drawing_difficulty": "Beginner",
    "skills": [
      "Basic object construction",
      "Numerical and metric constraints",
      "Natural language to tool mapping ability"
    ],
    "tasks": [
      {
        "function": "draw_polygon",
        "args": {
          "points": [[0,0], [2,0], [2,2], [0,2]]
        }
      }
    ]
  }
]
\end{lstlisting}

\textbf{Example Input 2}
\begin{itemize}[leftmargin=*, itemsep=1pt]
  \item \textbf{Question}: Construct triangle ABC and the angle bisector of angle A. Label the bisector as ``L1''.
  \item \textbf{Answer}: An angle bisector divides an angle into two equal angles.
  \item \textbf{Image}: Vertex A is on the left, base BC is horizontal.
\end{itemize}

\textbf{Example Output 2}
\begin{lstlisting}
[
  {
    "description": "Construct a triangle and its angle bisector labeled L1",
    "grade_level": "Grade 8",
    "drawing_difficulty": "Intermediate",
    "skills": [
      "Basic object construction",
      "Geometric relations and constraints",
      "Multi-step dependency planning ability",
      "Natural language to tool mapping ability"
    ],
    "tasks": [
      {
        "function": "draw_polygon",
        "args": {
          "points": [[-1,0], [3,0], [1,3]]
        }
      },
      {
        "function": "angle_bisector",
        "args": {
          "points": [[3,0], [-1,0], [1,3]]
        }
      },
      {
        "function": "add_text_label",
        "args": {
          "position": [0.5,1],
          "text": "L1"
        }
      }
    ]
  }
]
\end{lstlisting}

\vspace{2pt}
\textbf{8. Final Instructions}

Follow all rules above. Prioritize original-figure replication and output the mandatory JSON format.

\end{promptbox}

\captionof{figure}{Prompt used to generate structured GeoGebra construction tasks from K12 geometry problems. The model integrates question text, answer hints, and image context to infer construction steps, classify skills, assign grade level and drawing difficulty, and output a JSON-formatted task sequence.}
\label{fig:geogebra_task_generation_prompt}
\section{Dataset Quality Assurance Prompt}

\begin{promptbox}[Rigorous Multi-modal Dataset Quality Assurance Reviewer]
\sloppy\setlength{\emergencystretch}{1em}

You are a \textbf{Rigorous Multi-modal Dataset Quality Assurance Reviewer}. You will receive a dataset entry containing the following fields:

\begin{itemize}[leftmargin=*, itemsep=1pt]
  \item \textbf{question}: The text of the math problem.
  \item \textbf{answer}: The final answer without solution steps.
  \item \textbf{image}: The original image accompanying the problem, including visual information such as geometry, function graphs, or coordinate systems expected by the problem.
  \item \textbf{tasks}: A sequence of function calls executed in GeoGebra, where each step contains a function and args.
  \item \textbf{final\_canvas\_screenshot}: The final canvas screenshot obtained after executing the tasks on the GeoGebra website.
\end{itemize}

Your goal is to score the quality of this entry for dataset filtering purposes.

\vspace{2pt}
\textbf{1. Mandatory Review Process}

\textbf{A. Deconstruct Tasks Item by Item}

You must traverse strictly one by one through every task in \texttt{tasks}, treating each as an independent and verifiable requirement. Form a mental checklist of what object, curve, label, or relationship should correspond to each task on the final canvas.

\vspace{2pt}
\textbf{B. Visual Verification Item by Item}

For each task, perform visual verification on the \texttt{final\_canvas\_screenshot}. During verification and comparison, allow the screenshot to have a global rotation relative to the original \texttt{image}, such as 90 degrees, 180 degrees, 270 degrees, or other angles. As long as the relative positions and geometric relationships of key objects remain consistent after rotation, it should be considered basically restored or consistent. Do not judge it as inconsistent or lower the score solely due to rotation.

\vspace{2pt}
\textbf{2. Verification Criteria}

\textbf{A. Function or Curve Tasks}

For tasks involving function graphs, circles, lines, parabolas, piecewise functions, and other curves, check the following:

\begin{itemize}[leftmargin=*, itemsep=1pt]
  \item \textbf{Shape and Type}: Whether the curve type is correct, such as line, circle, parabola, polyline, or piecewise curve.
  \item \textbf{Key Feature Points}: Whether key points appear correctly and in reasonable positions, such as vertices, intercepts, centers, radii, endpoints, inflection points, or intersections.
  \item \textbf{Direction and Opening}: For example, whether a parabola opens upward or downward according to the expression.
  \item \textbf{Piecewise Correctness}: Whether the intervals of piecewise functions correspond to the correct segments or curves, and whether breakpoints or closed intervals are reasonably represented.
  \item \textbf{Relation to Coordinate System}: If the question or task involves coordinate axes or scales, the curve position should match the coordinate system. Minor scaling or translation is allowed if the mathematical properties remain correct.
\end{itemize}

\textbf{B. Point, Intersection, or Special Point Tasks}

For tasks involving plotted points, intersections, perpendicular feet, midpoints, and other special points, check the following:

\begin{itemize}[leftmargin=*, itemsep=1pt]
  \item \textbf{Existence}: Clearly visible point markers must exist.
  \item \textbf{Coordinates or Location}: The point should be located at the specified coordinates or geometric position. Small errors are allowed, but obvious deviation is not.
  \item \textbf{Dependency Relations}: If a point is supposed to be on a specific line, circle, or curve, it must actually fall on it.
\end{itemize}

\textbf{C. Label or Text Tasks}

For tasks such as \texttt{add\_text\_label}, check the following:

\begin{itemize}[leftmargin=*, itemsep=1pt]
  \item \textbf{Content Verbatim Consistency}: The label text must be exactly consistent with the text specified in \texttt{tasks}. Case sensitivity, math symbols, brackets, spaces, and units all count.
  \item \textbf{Positional Accuracy}: The label anchor point should be near the specified coordinates and clearly associated with the intended location.
\end{itemize}

\vspace{2pt}
\textbf{3. Scoring Tasks}

You must output three scores, each from 1 to 5, and one comprehensive result.

\begin{enumerate}[leftmargin=*, itemsep=1pt]
  \item \textbf{image Score}: Judge whether the \texttt{final\_canvas\_screenshot} basically restores the original \texttt{image}.
  \item \textbf{Auxiliary Line Score}: Judge whether the \texttt{tasks} and \texttt{final\_canvas\_screenshot} contain necessary and reasonable auxiliary elements surrounding the \texttt{question}, such as auxiliary lines, perpendiculars, intersections, annotations, coordinate axes, function curves, or unit labels.
  \item \textbf{tasks Score}: Based on task deconstruction and visual verification, judge whether the \texttt{final\_canvas\_screenshot} completely executes and presents the content required by \texttt{tasks}.
\end{enumerate}

If the sum of the three scores is \(\geq 9\), the comprehensive result is \texttt{True}; otherwise, it is \texttt{False}.

\vspace{2pt}
\textbf{4. Scoring Rubric}

\textbf{A. image Score: Consistency with Original Image}

\begin{itemize}[leftmargin=*, itemsep=1pt]
  \item \textbf{5 points}: Key objects are complete; geometric relationships and structures are highly consistent; subjects correspond at a glance.
  \item \textbf{4 points}: Subjects are consistent with minor non-critical differences, such as slight position deviation, scaling, line thickness, or font size.
  \item \textbf{3 points}: Partially consistent but with obvious missing parts or deviations.
  \item \textbf{2 points}: Only a few elements are similar; the overall structure is difficult to correspond.
  \item \textbf{1 point}: Basically irrelevant or obviously wrong.
\end{itemize}

\textbf{B. Auxiliary Line Score: Necessary Mathematical Constructions}

Auxiliary constructions include not only geometric segments, but also perpendiculars, parallel lines, extensions, intersections, perpendicular feet, midpoints, circles, angle annotations, coordinate axes, unit labels, and function graphs. The key criterion is whether they are useful for understanding or solving the problem.

\begin{itemize}[leftmargin=*, itemsep=1pt]
  \item \textbf{5 points}: Constructions fit the \texttt{question} highly; key relationships are expressed completely; no excessive clutter.
  \item \textbf{4 points}: Generally relevant and basically sufficient, with only minor missing or harmless extra elements.
  \item \textbf{3 points}: Relevant constructions exist, but key auxiliaries are missing or expression is not clear enough.
  \item \textbf{2 points}: Auxiliary constructions are mostly irrelevant or key auxiliaries are severely missing.
  \item \textbf{1 point}: Almost no construction around the problem; unable to assist understanding or solving.
\end{itemize}

\textbf{C. tasks Score: Correct Implementation of Tasks}

This score must be based on task deconstruction and visual verification, considering mathematical correctness, object type, key features, text consistency, and positional relationships.

\begin{itemize}[leftmargin=*, itemsep=1pt]
  \item \textbf{5 points}: Almost all tasks can be found in the canvas; function or shape types and key features are correct; text is exactly consistent; points and relations are correct.
  \item \textbf{4 points}: Most tasks are correctly presented, with only minor omissions or slight position deviations.
  \item \textbf{3 points}: Multiple tasks are missed or obvious inconsistencies exist.
  \item \textbf{2 points}: Only a small part of tasks is presented, with severe omissions or wrong key mathematical properties.
  \item \textbf{1 point}: Almost unexecuted or basically unrelated to \texttt{tasks}.
\end{itemize}

\vspace{2pt}
\textbf{5. Tolerance and Penalty Principles}

\begin{itemize}[leftmargin=*, itemsep=1pt]
  \item \textbf{Allow Reasonable Deviations}: Global scaling, translation, rotation, line thickness, color, and font size differences are allowed as long as mathematical relationships and recognizability are not altered.
  \item \textbf{Must Penalize Significantly}: Missing subjects, wrong geometric relationships, wrong shape types, missing objects explicitly required by \texttt{tasks}, and label text inconsistency must be penalized.
\end{itemize}

\vspace{2pt}
\textbf{6. Output Format}

Your response must be a JSON object with the following structure. Do not include explanations.

{\footnotesize\ttfamily
\noindent \{\\
\hspace*{1em}"image Score": "Score",\\
\hspace*{1em}"Auxiliary Line Score": "Score",\\
\hspace*{1em}"tasks Score": "Score",\\
\hspace*{1em}"Comprehensive Result": "True/False"\\
\}
}

\end{promptbox}

\captionof{figure}{Prompt used for multimodal quality assurance of GeoGebra-based dataset entries. The evaluator checks the original image, generated canvas, auxiliary constructions, and task execution quality, then outputs three scalar scores and a comprehensive filtering result.}
\label{fig:dataset_quality_assurance_prompt}

\section{Metrics}
\label{metrics}
To comprehensively evaluate a model's geometric generative reasoning capability, we design a four-stage evaluation protocol covering the middle process of action execution, the final geometric result, a stratified analysis, and a unified overall score.

\subsection{Middle Process Metrics (Rule-Based)}

To measure the quality of the step-by-step action sequence produced by the agent, we define four fine-grained rule-based metrics computed directly from the predicted action log against the ground-truth annotation.

Let the predicted action sequence be $\hat{A} = \{\hat{a}_1, \hat{a}_2, \ldots, \hat{a}_m\}$ and the ground-truth sequence be $A^* = \{a_1^*, a_2^*, \ldots, a_n^*\}$, where each step $a_i = (\text{type}_i, \text{param}_i)$.

\textbf{Action Type Accuracy (AA, Action)} measures per-step correctness of the predicted action category:
\begin{equation}
    \text{AA} = \frac{1}{|A^*|} \sum_{i=1}^{|A^*|} \mathbf{1}[\hat{\text{type}}_i = \text{type}_i^*]
\end{equation}

\textbf{Parameter Accuracy (PA, Param)} evaluates whether each predicted action's parameters are correct. For \texttt{type} actions, string equality (case-insensitive) is required. For \texttt{click} actions, the predicted coordinate $(\hat{x}, \hat{y})$ must fall within the annotated bounding box:
\begin{equation}
    \text{correct}_\text{click} = \mathbf{1}[x_\text{min} \leq \hat{x} \leq x_\text{max}\ \wedge\ y_\text{min} \leq \hat{y} \leq y_\text{max}]
\end{equation}
For \texttt{paint} actions, correctness is determined by the Euclidean pixel distance on the $1280 \times 720$ screen being within a 5-pixel tolerance:
\begin{equation}
    \text{correct}_\text{paint} = \mathbf{1}\!\left[\sqrt{(1280(\hat{x} - x^*))^2 + (720(\hat{y} - y^*))^2} \leq 5\right]
\end{equation}
The per-step correctness indicators are aggregated as:
\begin{equation}
    \text{PA} = \frac{1}{|A^*|} \sum_{i=1}^{|A^*|} \mathbf{1}[\text{param}_i\ \text{correct}]
\end{equation}

\textbf{Step Success Rate (SSR, Step)} requires both action type and parameter to be simultaneously correct at each step:
\begin{equation}
    \text{SSR} = \frac{1}{|A^*|} \sum_{i=1}^{|A^*|} \mathbf{1}[\hat{\text{type}}_i = \text{type}_i^*\ \wedge\ \hat{\text{param}}_i\ \text{correct}]
\end{equation}

\textbf{Task Success Rate (TSR, Task)} is a binary indicator per task requiring all steps to be simultaneously successful, averaged across tasks:
\begin{equation}
    \text{TSR} = \frac{1}{T}\sum_{t=1}^{T}\mathbf{1}\!\left[\forall\, i \in t:\ \hat{\text{type}}_i = \text{type}_i^*\ \wedge\ \hat{\text{param}}_i\ \text{correct}\right]
\end{equation}
where $T$ is the total number of tasks.

All four metrics are first computed per task, then averaged across tasks within each evaluation ID, and finally macro-averaged across all IDs to obtain the model-level score. The \textbf{Middle Process Score (MPS, Middle)} is then computed as a weighted combination:
\begin{equation}
    \text{MPS} = 0.6 \cdot \text{TSR} + 0.2 \cdot \text{SSR} + 0.1 \cdot \text{PA} + 0.1 \cdot \text{AA}
\end{equation}
The weights reflect the hierarchy of difficulty: task-level success is the strictest criterion and thus contributes most, while action type accuracy serves as a lower-level sanity check.

\subsection{Final Result Metrics}

We evaluate the quality of the model's final geometric output from two complementary perspectives.

\paragraph{Objective Task Completion Score (OTC, O-Comp.).}
Given the ground-truth GeoGebra construction $G^* = (\mathcal{P}^*, \mathcal{C}^*)$ and the model's construction $\hat{G} = (\hat{\mathcal{P}}, \hat{\mathcal{C}})$, where $\mathcal{P}$ denotes the set of labeled points with coordinates and $\mathcal{C}$ denotes the list of construction commands, we compute a geometric similarity score. For each model point $\hat{p} \in \hat{\mathcal{P}}$, we find its closest ground-truth match and define a continuous point-match contribution via an exponential kernel:
\begin{equation}
    s_\text{point} = \frac{1}{|\mathcal{P}^*|} \sum_{\hat{p} \in \hat{\mathcal{P}}} \exp\!\left(-5 \cdot d_{\min}(\hat{p}, \mathcal{P}^*)\right) \cdot \mathbf{1}[d_{\min} \leq \tau]
\end{equation}
where $\tau = 0.5$ is the normalized coordinate tolerance. For the command sequence, we translate each model command's inputs through the point correspondence mapping and check for a signature match $(\text{cmd\_name},\ \text{sorted\_inputs})$ against the ground-truth command set:
\begin{equation}
    s_\text{cmd} = \frac{|\hat{\mathcal{C}}_\text{matched}|}{|\mathcal{C}^*|}
\end{equation}
The task completion score combines both components, with geometric structure weighted more heavily than point placement:
\begin{equation}
    \text{OTC} = 0.4 \cdot s_\text{point} + 0.6 \cdot s_\text{cmd}
\end{equation}

\paragraph{VLM-Based Holistic Scores.}
To capture aspects that resist purely symbolic evaluation---such as visual rendering fidelity, label legibility, and cross-modal consistency---we employ Gemini-2.5-Pro as an automated evaluator. Given the task description, ground-truth JSON construction data, model JSON data, ground-truth rendered image, and model rendered image, the VLM outputs scores in $[0, 1]$ across two evaluation views.

Under the \textit{data logic} view and the \textit{visual presentation} view, the model independently scores three dimensions:
\begin{itemize}
    \item \textbf{Task Completion (TC, S-Comp.)}: whether all geometric elements defined in the task list are present and identifiable in the output.
    \item \textbf{Visual Similarity (VS, S-Vis.)}: pixel-level rendering fidelity relative to the ground-truth image, including line thickness, color, label positioning, and coordinate scaling.
    \item \textbf{Geometric Logic (GL, S-Geo.)}: mathematical correctness of geometric relationships such as perpendicularity, tangency, and midpoints, including detection of visual ``hallucinations'' where the image appears correct but the underlying coordinates are logically inconsistent.
\end{itemize}
An \textit{overall comprehensive} view synthesizes the scores from both the \textit{data logic} view and the \textit{visual presentation} view into a final set of scores $(\text{TC}, \text{VS}, \text{GL})$, where each dimension is the average of its counterparts across the two views.

The \textbf{Final Result Score (FRS, Final)} is a weighted combination of the rule-based and VLM-based scores:
\begin{equation}
    \text{FRS} = 0.3 \cdot \text{OTC} + 0.3 \cdot \text{TC} + 0.2 \cdot \text{VS} + 0.2 \cdot \text{GL}
\end{equation}

\subsection{Overall Score}

To facilitate model comparison with a single scalar, we define the \textbf{Overall Score (OS, Overall)} as the equally-weighted average of MPS and FRS:
\begin{equation}
    \text{OS} = \frac{1}{2}\left(\text{MPS} + \text{FRS}\right)
\end{equation}
The overall score thus balances procedural correctness in the action sequence against the quality of the final geometric output, providing a unified ranking across all evaluated models.

\newpage

\end{document}